\documentclass{article}

\usepackage[utf8]{inputenc}
\usepackage[T1]{fontenc}
\usepackage{geometry}
\usepackage[dvipsnames,table]{xcolor}
\definecolor{systemblue}{RGB}{52,152,219}
\usepackage{graphicx}
\usepackage{amsmath,amssymb}
\usepackage{booktabs,array,longtable}
\usepackage{fancyhdr,float}
\usepackage{caption,subcaption}
\usepackage{natbib}
\usepackage{textgreek}
\usepackage[inline]{enumitem} 
\usepackage{nameref}
\usepackage{multirow}
\usepackage{amsfonts}  
\usepackage{dsfont}
\usepackage[ruled,vlined]{algorithm2e}
\usepackage{listings}
\usepackage{tcolorbox}
\tcbuselibrary{listings}
\usepackage[title]{appendix}
\usepackage{threeparttable}
\usepackage{tikz}
\usepackage{pgfplots}
\pgfplotsset{compat=1.18}
\usepackage{csquotes}
\usepackage[utf8]{inputenc}
\DeclareUnicodeCharacter{202F}{\,}

\SetKwProg{Fn}{Function}{}{end Function}
\let\Fn\Fn
\SetKwProg{Try}{try}{}{end try}
\SetKwProg{Catch}{catch}{}{end catch}
\SetKw{Return}{return}
\SetKwComment{tcc}{// }{}  
\SetKwComment{tcp}{// }{}
\usepackage[
colorlinks=true,
linkcolor=systemblue,
urlcolor=systemblue,
citecolor=systemblue,
pdftitle={Self-evolving expertise in complex non-verifiable domains: dialogue as implicit meta-RL},
pdfauthor={Richard M. Bailey}
]{hyperref}

\title{\textbf{Self-evolving expertise in complex non-verifiable subject domains: dialogue as implicit meta-RL}}
\author{Richard M. Bailey, 
University of Oxford\\
\texttt{\href{mailto:richard.bailey@ouce.ox.ac.uk}{richard.bailey@ouce.ox.ac.uk}}}
	
\begin{document}
	
	\maketitle
	
	\begin{abstract}
		So-called \emph{wicked problems}, those involving complex multi-dimensional settings, non‑verifiable outcomes, heterogeneous impacts and a lack of single objectively “correct” answers, have plagued humans throughout history. Modern examples include decisions over justice frameworks, solving environmental pollution, planning for pandemic resilience and food security.
		The use of state-of-the-art artificial intelligence systems (notably Large Languag Model-based agents) `collaborating' with humans on solving such problems is being actively explored.
		While the abilities of LLMs can be improved by, for example, fine-tuning, hand‑crafted system prompts and scaffolding with external tools, LLMs lack endogenous mechanisms to \emph{develop expertise through experience} in such settings. 
		This work address this gap with \emph{Dialectica}, a framework where agents engage in structured dialogue on defined topics, augmented by memory, self‑reflection, and policy‑constrained context editing. Formally, discussion is viewed as an \emph{implicit meta‑reinforcement learning} process. 
		The `dialogue-trained' agents are evaluated \textit{post-hoc} using judged pairwise comparisons of elicited responses. Across two model architectures (locally run \textit{Qwen3:30b} and OpenAI’s \textit{o4‑mini}) results show that enabling reflection-based context-editing during discussion produces agents which dominate their baseline counterparts on Elo scores, normalized Bradley–Terry–Davidson ability, and AlphaRank mass.
		The predicted signatures of learning are observed qualitatively in \textit{statement} and \textit{reflection} logs, where reflections identify weaknesses and reliably shape subsequent statements. Agreement between quantitative and qualitative evidence supports dialogue‑driven context evolution as a practical path to targeted expertise amplification in open non-verifiable domains. \\
		
	\end{abstract}
	
	\vspace{2em}
	
	\begin{figure}[h]
		\centering
		\begin{minipage}[t]{0.7\textwidth} 
			\vspace{0pt}
			\includegraphics[width=\textwidth]{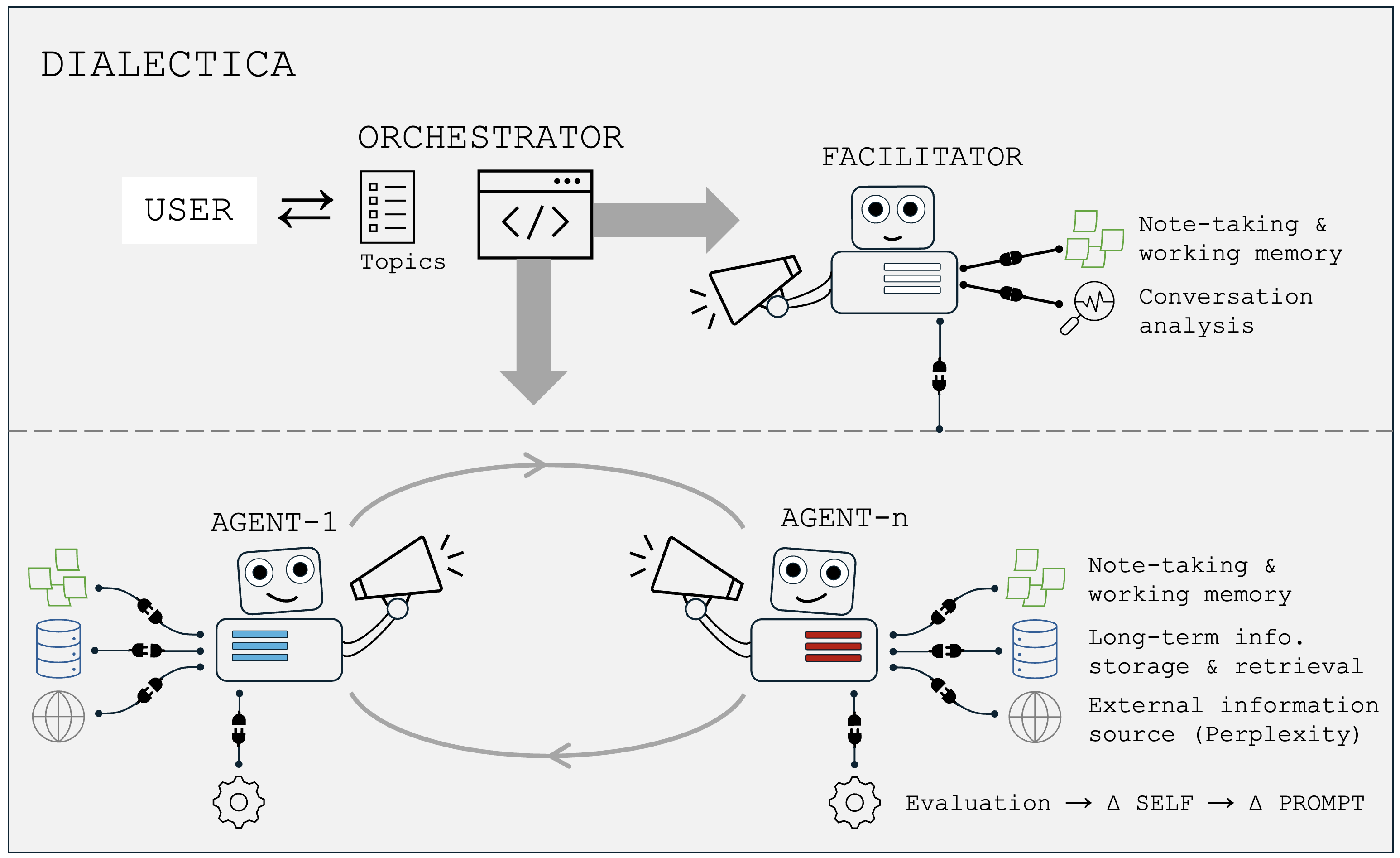}
		\end{minipage}
		\hfill
		\begin{minipage}[t]{0.27\textwidth}  
			{\footnotesize
				\caption*{\small
					The main components of the system: the Orchestrator is a deterministic
					component responsible for calling all actions and holding the list of discussion topics;
					the agents take turns to deliver statements, which are broadcast to all other agents.
					Agents have configurable use of tools for working memory, external information collection,
					storage and retrieval, which they use to update their prompt context over time.
					The (optional) Facilitator helps to guide the discussion.%
				}\label{fig:figure1}}
		\end{minipage}
	\end{figure}

	\newpage
	\tableofcontents
	\newpage

	\section{Introduction}
	Many of the important problems facing society are so-called \textit{wicked problems} \citet{rittel1973dilemmas} which exist in complex multi-dimensional settings, have no simple objectively correct answers, where tradeoffs occur, legitimate opposing positions can be held, information heterogeneity exists, and where the range of possible solutions heterogeneously affect different stakeholders. Examples include the design of justice frameworks, solving environmental pollution, planning for pandemic resilience, global food security. These are also non-verifiable domains in the sense there are no generally agree-upon metrics to measure proximity to \emph{correct opinions} or \emph{optimal outcomes}. Applying powerful artificial intelligence (AI) tools to such problems requires new approaches, and the thesis put forward here is that \emph{agent-to-agent dialogue} is helpful in this regard. Dialogue has emerged in human society as a way of collectively searching for solutions to difficult problems.
	This idea has a long history. For example in Western thought, from Socrates’ dialectical questioning aimed at clarifying truth, through Aristotle’s logical systemization, Cicero’s presentation of opposing arguments, to Enlightenment thinkers such as Kant and Hume, who saw critical public reasoning and skeptical dialogues as key to intellectual advancement. Modern philosophers such as Habermas and Popper have further emphasized debate as central to knowledge formation, legitimacy, and societal openness. 
	Such interactions take many forms, including negotiations, parliamentary process, structured meetings between individuals and groups, and everyday conversation. While the essential goals of dialogue, discussion and debate are in some sense clear (identify weaknesses in ideas, test the scope of solutions, reach concensus, for example) the unavoidable biases and limitations of human reasoning puts fundamental limitations on our problem-solving capabilities. This present work is an attempt to distill the rich utility of discussion and debate in to a simplified process for improving Large Language Model (LLM) agents' ability to aid humans in tackling complex open-ended questions. It is not intended to be a simulation of human debates but a process to produce AI entities which have `benefitted' from discussion and are able to produce more sophisticated outputs and therefore be more useful ongoing collaborators than their baseline counterparts. 
	
	This work builds on a rich recent history of research on self-improving LLM agents, which can be organised by mechanism and operational domain, as follows. Adaptive coding agents e.g., the Adaptive Self-Improvement framework for ML code by \citet{zhang2025adaptiveselfimprovementllmagentic} and SICA by \citet{robeyns2025selfimprovingcodingagent}, iterate candidate solutions, analyze failures, and revise prompts/tools in verifiable settings (code benchmarks, engineering tasks). Meta-learning and fine-tuning methods, SEAL \citep{zweiger2025selfadaptinglanguagemodels} and SiriuS \citep{zhao2025siriusselfimprovingmultiagentsystems}, self-generate training data or distill successful reasoning traces across QA, cooperative reasoning, and negotiation, spanning moderately verifiable to more open-ended domains. Self-play reinforcement learning, RAGEN \citep{wang2025ragenunderstandingselfevolutionllm} and the Diplomacy agent Richelieu \citep{guan2024richelieuselfevolvingllmbasedagents}, refine strategies via repeated play on puzzles and strategic games, with improvements measured by scores/win rates. Finally, general-purpose self-referential agents, the Gödel Agent \citep{yin2025godelagentselfreferentialagent} and Agents of Change \citep{belle2025agentschangeselfevolvingllm}, rewrite internal logic or policies across tasks from puzzles to multi-player games. Collectively, these approaches span a spectrum from highly structured, verifiable improvements (games, code) to increasingly open-ended interactions, highlighting substantial promise in self-improving LLMs.
	In parallel, forced debates between LLMs have been explored for improving reasoning and evaluation. Three strands can be resolved: (i) \emph{training-based} self-play/adversarial debate to enhance reasoning via feedback \citep{arnesen2024traininglanguagemodelswin, chen2024selfplayfinetuningconvertsweak, cheng2025selfplayingadversariallanguagegame, kirchner2024proververifiergamesimprovelegibility}; (ii) \emph{evaluation frameworks} where debate assists a judge on QA/reading-comprehension benchmarks \citep{khan2024debatingpersuasivellmsleads, kenton2024scalableoversightweakllms, moniri-etal-2025-evaluating, chan2023chatevalbetterllmbasedevaluators}; and (iii) \emph{constitutional/self-critical} approaches that guide internal debate by principles to promote truthfulness and safety \citep{bai2022constitutionalaiharmlessnessai}. These methods typically rely on adversarial setups with binary outcomes and judge adjudication, making them vulnerable to rhetorical optimization, judge bias, obfuscation, and collusion \citep{koo2024benchmarkingcognitivebiaseslarge}. Current questions include extending debate to nuanced, open-ended domains, ensuring judges reward truthfulness rather than persuasive rhetoric, and preventing gaming or collusion among capable debaters \citep{browncohen2025avoidingobfuscationproverestimatordebate, kenton2024scalableoversightweakllms}.

	\subsection{This contribution}
	The goal of this work is to produce entities (AI `collaborators') which have sophisticated robust views on the `wicked problems' we (humans) find difficult to solve. 
	The logic of this work starts with the simple well-known observation that \textit{system prompts} are important in determining model outputs. As an extension, the \emph{context} (the information contained within the prompt in addition to the direct \emph{question} being asked) is also important in determining model outputs, an example being the abilities of models to perform `in-context learning'. The natural question then is how can we optimise what is in the context in orer to get the best responses. In the present case we are focusing on problems for which we cannot explicitly score the answers produced by the LLM. Therefore we cannot use formal optimization, as we have no objective function. The research question then becomes that of finding a way to update the prompt context which is consistent with improved LLM reponses to these `non-scorable wicked problems'.
	
	The hypothesis being tested is that sending LLM agents through multiple discussions, each with multiple rounds, allowing them to freely gather information, argue, reflect and evolve their views, is equivalent to an implicit form of learning which will result in more nuanced arguments and generally better-informed agents.
	The \emph{Dialectica} system differs from existing self-evolving LLM agent frameworks in several respects. While previous work has concentrated on self-improvement in domains with clear success metrics (such as coding correctness \citep{zhang2025adaptiveselfimprovementllmagentic,robeyns2025selfimprovingcodingagent}, game outcomes \cite{wang2025ragenunderstandingselfevolutionllm,guan2024richelieuselfevolvingllmbasedagents}, or QA accuracy \citep{zweiger2025selfadaptinglanguagemodels,zhao2025siriusselfimprovingmultiagentsystems}) this present system operates in contested domains where multiple valid perspectives exist. 
	The approach is to use LLM agents in a semi-adversarial setting in the sense that agents have different initial world views, but are not explicitly trying to \emph{beat} each other. Importantly, agents are not given an explicit debate purpose (e.g., \textit{convince others, find consensus, explore the topic}), nor any success criteria or winning conditions. There is no goal orientation beyond staying true to their worldview, and no competitive or collaborative framing is provided. The system allows agents to develop their own understanding of what constitutes effective discussion participation based on their worldview and priorities, learning from opponent responses, and reflection on what strategies work. This open-ended approach means any goals which emerge do so endogenously through reflection and the context evolution process, as described below. As the agents are not given a purpose of \textit{winning the debate}, there are no incentives to develop unhelpful adversarial strategies such as elaborate rhetoric or bluffing, to persuade a judge (AI or human). Success therefore cannot be measured through simple performance metrics, there are no `winners and losers'. Rather than `optimizing' for victory through competitive interactions or self-play, agents in this system engage with opponents who (at least at first) hold different worldviews, learning through structured dialogue rather than strategic gaming. The system uses LLM-powered reflection to consolidate experiences across multiple discussion sessions, allowing agents to develop evolving perspectives (through accumulated experience and reflection; with no model fine-tuning) to reconstitute their prompt contexts. Following the discussion period, to judge which, if any, features of their experience led to improved abilities, individual agents were asked a common set of questions, and their outputs judged in pairwise comparisons.
	
	To best knowledge, this is the first system in which multiple LLM agents adjust their prompt context \emph{online}, using nothing but the unlabelled conversational signals exchanged between themselves, thereby instantiating a form of \emph{implicit meta-reinforcement learning} for non-verifiable tasks.
	
	\subsection{Dialogue as meta-learning}
	\label{meta_learning}
	In this paper, “meta-learning” (through dialogue) denotes an entity’s ability to acquire reusable rules, habits, scaffolds, and relevant information so that later responses are better structured and more sophisticated. The hypothesis being tested is that, as feedback during conversation provides these learning opportunites for humans, can do the same for LLM agents. through purposeful changes to the content of the model’s context, driven by conversational feedback and reflection (essentially pushing the model’s hidden state toward a preferred region of its activation manifold). Even without an explicit scalar reward, the signals from conversation could potentially drive an optimization over contexts, exploiting the model’s in‑context adaptation. 
	
	More formally, if the model’s output distribution \(p_\theta(y\mid c,x)\) depends smoothly on the context \(c\) (via its serialization and embedding), then systematically editing \(c\) based on conversational feedback is a form of \emph{conditioning control}. 
	In the absence of an explicit numeric reward, conversation provides \emph{implicit signals} (criticisms, endorsements, contradictions, and the agent’s own reflections) that indicate directional pressure. Iterating this yields a black‑box optimization over contexts (even as the objective remains hidden both from the external observer and the participants).
	A parsimonious way to view conversational turns, interspersed with periods of reflection, is as a two‑time‑scale learning process with fixed model parameters \(\theta\). At the \emph{interaction} (inner) time scale, an agent produces text outputs by sampling from a context‑ and guidance‑conditioned policy
	\[
	p_\theta(y\mid x_t, C_t, P_k) \;\equiv\; \pi_\theta(\cdot\mid x_t,C_t,P_k),
	\]
	where \(x_t\) is the current task/input at turn \(t\); \(C_t\) is the serialized dialogue context; and \(P_k\) is a vector of \emph{guidance variables} (e.g., instructions, core beliefs, style/strategy priors, tool‑use preferences) that change on a slower adaptation time index \(k\). 
	Qualitative cues from the exchange (statements of other agents, reflections) are retained as internal traces.
	At the \emph{adaptation} (outer) time scale, a meta‑policy \(\Pi_{\psi}\) maps a compressed state \(S_k \;=\; g\!\big(C_{1:T_k},\, P_k,\, \text{traces}\big)\)
	to a bounded update of the guidance variables,
	\[
	\Delta P_k \;\sim\; \Pi_{\psi}(\cdot \mid S_k),
	\qquad
	P_{k+1} \;=\; P_k \oplus \Delta P_k,
	\]
	where \(g(\cdot)\) summarizes recent turns and reflections; \(\oplus\) is a policy‑constrained operator that clips, validates, and maps the proposal into the admissible set (ensuring the stability and safety of updates); and \(T_k\) indexes the last interaction turn included in the \(k\)-th summary. The outer‑loop policy \(\Pi_{\psi}\), given the summarized meta‑state \(S_k\), proposes a (possibly stochastic) edit \(\Delta P_k\) to the guidance variables. After projection via \(\oplus\), this edit updates the \emph{conditioning state} (the context and guidance variables that seed the prompt) used by the inner policy \(\pi_\theta\) on subsequent turns. The meta‑parameters \(\psi\) are distinct from the model weights \(\theta\) used by \(\pi_\theta\); in practice, \(\psi\) includes the outer‑loop prompt templates (reflection/edit instructions), decoding hyperparameters, the allow‑list and guardrail settings (which prompt components are editable), and any thresholds used by \(\oplus\). We do not learn \(\psi\) in these experiments; it is fixed by the experimental condition.
	Because conversational partners may also adapt, each agent learns under non‑stationary feedback, which could be a source of instability; the bounded updates and compression \(g(\cdot)\) regularize this process to some extent. 
	Under mild regularity conditions (smooth dependence of \(p_\theta(\cdot\mid x, c)\) on \(c\) and bounded step sizes in \(\oplus\)), the outer loop acts as a self‑consistent update map on \(c_t\) that seeks a local \emph{reflective equilibrium} where edits diminish and statements align with guidance. 
	In this sense, conversation functions as an outer‑loop optimizer over contexts while the fixed model supplies inner in‑context adaptation, implementing a meta‑learning process in settings without explicit verifiable rewards.
		
	\subsection{Key findings}
	\label{key_findings}
	\begin{enumerate}[leftmargin=0.75cm]
		\small
		\item \textbf{Dialogue-driven context evolution significantly improves performance.}
		Structured dialogue with reflection and context-editing significantly enhanced agent capabilites in all open non-verifiable domains tested. Across three tournaments and two model families, self-reflecting agents that evolve their context consistently dominated non-evolving agents on all ranking signals reported. All major improvements arose from updates to the conditioning state $c_t=(C_t,P_k)$, demonstrating that evolution of the prompt context is the primary driver.
		
		\item \textbf{Micro‑level signatures of implicit meta‑learning are observable in discussion records.}
		Records of agent internal reflections and subsequent statements show clear consistent patterns consistent with small, regularized shifts in $p_\theta(\cdot\mid x_t,c_t)$ induced by context edits: uptake of novel ideas (reflections $\rightarrow$ next statements), systematic response (e.g., introduction of relavent concepts/features) following critiques, and coherent turn‑to‑turn trajectories (fewer contradictions, richer justifications). 
		
		\item \textbf{Gains are model‑agnostic, with larger relative improvements for the locally‑run model.}
		Both Qwen3‑30b (local) and OpenAI’s o4‑mini (proprietary) deliver improved responses following `conversational training'. Qwen3‑30b exhibits particularly large relative gains, approaching or surpassing the baseline performance of the stronger o4‑mini model, while o4‑mini also improves but more modestly. Pooled analyses show \emph{o4‑mini} (with memoriy, context evolution and web tools enabled) powers the top-performing agent.

	\end{enumerate}

	\section{System description}
	\subsection{Over-view}
\begin{figure}[h]
		\centering
		\includegraphics[width=0.8\textwidth]{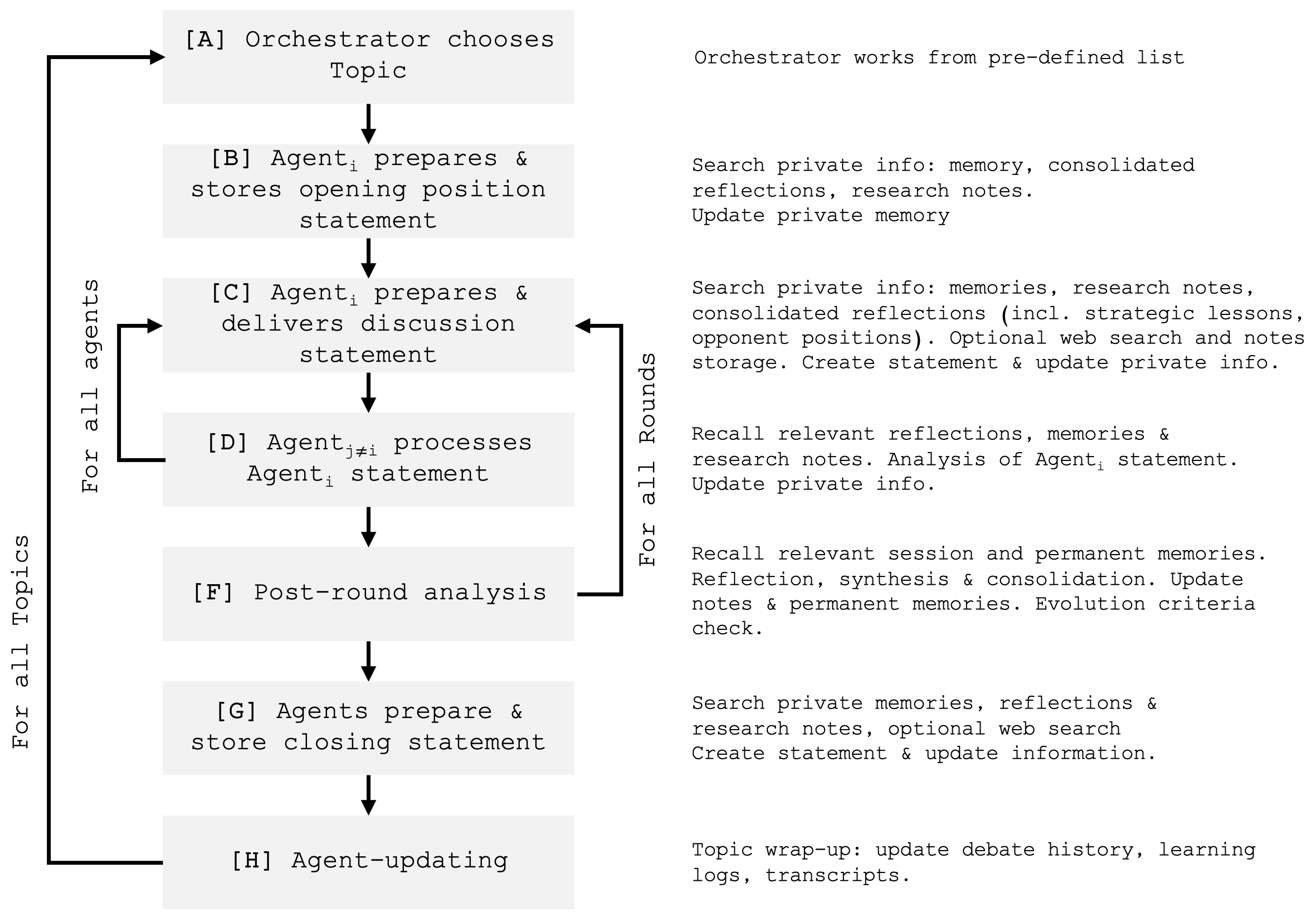} 
		\caption{In each discussion experiment, a set of \emph{Topics} are discussed over several \emph{Rounds} of \emph{Orchestrator}-guided interactions between multiple \emph{Agents}. During each round of dialogue, agents use memory and other tools to gather information and have the ability to edit the information used to generate their prompt context prior to their next conversation statement.}
		\label{fig:discussion_flow}
	\end{figure}

	Each agent is a specified persona, augmented by a variety of tools including memory and the ability to perform web searches.  
	A pre-defined list of \emph{Topics} are discussed by a fixed set of such agents over multiple \emph{Rounds} of dialogue.
	When a discussion starts, an \emph{Orchestrator} loads the agent’s configuration, enables its working-memory tool, and toggles optional RAG, web search and context-editing tools. 
	For each round of discussion 
	the orchestrator delivers key information to each agent including discussion topic, round index, accumulated arguments, current positions, and an ongoing transcript). Agents are called on to make statements in response to the topic in question and the statements of other agents. 
	Each time the agent produces a statement, its prompt is rebuilt from four components: (i) its identity configuration object (description, worldview, priorities, debate and communication style, preferred evidence types); (ii) a strict evidence-only instruction; (iii) snippets of text returned by the composite retrieval tool (e.g. recent reflections, consolidated lessons, transcript records, tracked opponent claims); and (iv) evidence gathered from RAG or web search tools. 
	After each round of the topic discussion (after all agents have spoken), the system captures (LLM-generated) personal reflections from each agent, and agents performs topic-level consolidation across all reflections. When \emph{evolution criteria} are met, these reflections trigger controlled updates to each agent's configuration. When discussion of the \emph{topic} concludes, agents record closing syntheses, append a session summary to the reflection-based debate history, and persist transcripts for future retrieval.
	The progression of actions is indicated in Figure ~\ref{fig:discussion_flow}.
	In the present experiments we use the Qwen3-30b model served locally via Ollama and OpenAI’s o4-mini model via API. Agent communication and tool calling are implemented using Google's \emph{A2A} framework and Anthropic's Model Context Protocol respectively. 
	
	\paragraph{Agent basics}
	We model agent \(i\) engaged in a discussion at round \(r\in\mathbb{N}\) by
	\[
	A_{i,r}\;=\;\big(S_{i,r},\,M_{i,r},\,\tau_i\big),
	\]
	where \(S_{i,r}\) is the configuration state used to build the identity block within the prompt context, \(M_{i,r}\) is the memory component, and \(\tau_i\) encodes tool availability and policy flags (e.g., \emph{RAG documents system}, \emph{web‑search}, \emph{context evolution}). 
	During the discussion process, prompts $\Phi_i(r)$ are created from the agent's current configuration, retireved stored information, discussion round context, and tool flags. The LLM produces the utterance/statement conditioned on that prompt, ($u_r \sim \pi_\theta(\cdot \mid \Phi_i(r))$) and these statements are broadcast to the other agents in the conversation. 
	
	Agent identities are defined declaratively in per‑agent files and loaded at runtime into strongly typed configuration objects that drive both the agent’s A2A “agent card” and the identity block, $S_{i,r}$, which collects the agent’s configuration fields: natural language description, worldview, priorities, debate style, communication style, expertise domains, preferred evidence types, evolution settings, and metadata.
	The system rebuilds prompts from this configuration on every turn, so identity deterministically shapes rhetorical posture and information use at generation time. As described in more detail below, identity preservation is enforced by a policy that partitions fields into an allow‑list of evolvable components and a protected set. Examples of initial agent definitions, and of final definitions following updates during the debate process, are provided in the Appendix.

	\paragraph{Prompt Construction}
	On every turn, the system assembles the prompt context by constructing the agent identity block from configuration $S_{i,r}$ and supplementing it with retrieved information including in-memory notes, persistent notes and learning-log excerpts, returning a bounded set of “lessons” and “insights” inserted as structured sections in the prompt.
	The information available for factual claims at round $r$ is thus
	\[
	I_i(r) \;=\; K_{\text{rag}}(q_r)\;\cup\;K_{\text{web}}(s_r)\;\cup\;\mathcal{R}\!\big(q_r, M_{i,r}\big),
	\]
	where $q_r$ is a retrieval query (topic- or evidence-driven), $s_r$ is a web-search query (if permitted), and $\mathcal{R}(q_r,M_{i,r})$ denotes memory retrieval conditioned on the current memory state $M_{i,r}$. 
	During the conversation rounds, further relevant context is defined as
	\[
	\mathrm{ctx}_r \;=\; \big(T_r,\,\mathcal{A}_{<r},\,\mathcal{O}_r,\,\mathcal{D}_r\big),
	\]
	with \(T_r\) the current conversation topic, \(\mathcal{A}_{<r}\) previous‑round arguments, \(\mathcal{O}_r\) tracked opponent claims for this turn, and \(\mathcal{D}_r\) the instruction associated with the agent’s persona. 
	The dynamic, memory-conditioned prompt is then
	\[
	\Phi_i(r) \;=\; f_{\text{prompt}}\!\big(S_{i,r},\, I_i(r),\, \mathrm{ctx}_r,\, \tau_i\big),
	\]
	where $\tau_i$ is the list of available tools.

	\subsection{Memory architecture}
	Each agent is provided with a dual‑tier memory system: a session tier that records ephemeral details of the current interaction, and a persistent tier that accumulates durable knowledge and learning signals across debates and topics. 
	Thus,
	\[
	M_i(t) \;=\; M_i^{\text{sess}}(t)\;\cup\; M_i^{\text{pers}}(t),
	\]
	where $M_i^{\text{sess}}$ contains session notes and $M_i^{\text{pers}}$ contains persistent notes. We write a memory note as $n \in M_i(t)$ with attributes
	\[
	c(n)\in \mathcal{X} \;\;(\text{text content}),\quad \mathrm{imp}(n)\in\{\textsf{high},\textsf{medium},\textsf{low}\},\quad \tau(n)\in\mathbb{R}_{\ge 0}\;(\text{timestamp}).
	\]
	
	The session tier holds a chronological session notes file, an opponent‑tracking ledger keyed by opponent identifiers, and a thinking log that records private “scratch‑pad” reasoning and meta‑observations as they occur. The persistent tier maintains a long‑lived notes file for high‑importance material, a personal reflections log with rotation (to bound growth), a consolidated topics file that stores LLM‑synthesized perspectives keyed by normalized topic identifiers, and a reflection‑based debate history that tracks aggregate participation, topic expertise, strategic lessons, and memorable sessions. 
	The categorical importance is mapped to a numeric weight $I(n)$ as $\{1.0,0.6,0.3\}$ as $\{high, medium, low\}$ respectively.
	Because the memory tool is exposed through the system’s MPC, every call (save, retrieve, consolidate, or evolve) passes through a single, instrumented interface, and its effects are visible for inspection.
	
	Selective retrieval from memory is achieved using semantic search. Let $\phi(\cdot)$ denote the encoder (here, all-MiniLM-L6-v2), $q$ the query and $n$ the note. Semantic relevance is defined as cosine similarity
	\[
	R(q,n) \;=\; \frac{\langle \phi(q),\, \phi(c(n))\rangle}{\|\phi(q)\|_2\,\|\phi(c(n))\|_2}\;\in\;[-1,1],
	\]
	which is well-behaved in the empirically-observed range $[0,1]$ for typical text. 
	An option exists to also include a recency score. Let $\Delta t(n)$ denote the age of $n$; the recency score $T(n)\in [0,1]$ is then a decaying function, or piecewise decay, of $\Delta t(n)$ . This functionality was not applied in the present case and we set the weighting to zero below.
	The composite retrieval score is then
	$\mathrm{Score}(n\mid q) \;=\; 0.6\,R(q,n)\;+\;0.4\,I(n)\;+\;0\,T(n)$,
	with a moderate relevance floor $R(q,n)\ge \theta_R$ (e.g., $\theta_R\approx 0.2$) to filter weak matches. The retrieval operator $\mathcal{R}(q, M_i)$ returns the top-$k$ notes by $\mathrm{Score}$, optionally drawing from both session and persistent files.

	\subsection{Learning Dynamics}
	\label{Learning_dynamics}
	The agent earning process operates at three tightly-coupled stages. First, every speaking turn rebuilds the prompt context from strategic notes, consolidated reflections, opponent tracking, and (when enabled) external evidence, so the agent adapts its argument round by round. Second, immediately after each round the orchestrator triggers a per-agent reflection that saves any fresh insights the agent has had, updates topic-level consolidated memory, and (if context evolution thresholds are satisfied) applies allowed configuration edits. Finally, when all rounds of discussion for the current topic are complete, agents run a session-end process which writes the debate to persistent history, logs new strategic lessons, and stores the transcript ready for future use.
	
	\paragraph{Post-round consolidation}
	After each round, the agent produces an LLM‑assisted, first‑person reflection based on the most recent exchanges. A reflection record \(u\in \mathcal{U}_{i,T_r}\) includes fields \(\,r\) (round index), \(T_r\) (topic), \(\text{argument-summary}\), \(\text{opponent-count}\), \(\text{free-text}\) (what worked/failed and planned adjustments), and \(\text{timestamp}\). Entries are appended to a rotating log, maintaining a bounded window of recent items (defaulting to the last 100), with older lines archived to timestamped files; a session scratchpad mirrors the same content.
	Topic‑level LLM-based consolidation activates when at least one prior reflection exists for the same topic. Let \(\kappa(\cdot)\) be the normalized topic key. Two topics \(T,T'\) are considered similar enough for consolidation if the Jaccard similarity of their de‑noised token sets exceeds a threshold \(\tau_J\):
	\[
	J(T,T') \;=\; \frac{\big|\,\mathcal{W}(T)\cap \mathcal{W}(T')\,\big|}{\big|\,\mathcal{W}(T)\cup \mathcal{W}(T')\,\big|}\;\ge\;\tau_J,\qquad \tau_J=0.5,
	\]
	where \(\mathcal{W}(\cdot)\) removes common function words and very short tokens. Given reflections \(\mathcal{U}_{i,\kappa}\) on key \(\kappa\), the consolidation operator \(\mathcal{C}\) produces a structured summary
	\[
	C_{i,\kappa} \;=\; \mathcal{C}\big(\mathcal{U}_{i,\kappa}\cup\{u_{\text{new}}\}\big)
	\;=\; \big(\text{consolidated-perspective},\, \text{key-insights},\, \text{strategic-learnings},\, \text{meta}\big).
	\]
	The consolidated object is written to file under the normalized topic key, logged as a learning event, and made available for subsequent prompt construction `high' signal \emph{Strategic lessons} that guide argument structure and emphasis.
	
	\paragraph{Context evolution}
	Where permitted by experimental conditions and \texttt{evolution-settings} in the agent configuration, the system can upgrade the agent’s persona in a controlled, criteria‑gated process. 
	
	Because $\Phi_i(r)$ is rebuilt from $S_{i,r}$ on each turn, $S_{i,r}^{+}$ immediately and durably affects future prompts, while protected components in $\mathcal{P}$ preserve core identity (e.g., expertise domains; fundamental perspective or communication-style dimensions marked non-editable).
	
	After consolidation, the agent evaluates eligibility against default thresholds (production defaults: at least three debates and two consolidated topics; configurable via experiment scripts). If eligible, the system constructs an evolution prompt that includes: the current configuration (description, perspective, priorities, debate-style, communication-style, preferred evidence types); the evolution policy (intensity and which components are evolvable or protected); the latest consolidated perspective with its key insights and strategic learnings; and a compact survey of consolidated learning across topics. The LLM determines whether evolution should occur and, if so, returns a structured change proposal. The update is applied mechanically to the agent’s JSON configuration file, strictly limited to the allow‑listed components declared in evolution-settings.
	
	More formally, context editing is gated by experience thresholds and a policy over editable fields. Let $D_i$ be total debates completed and $U_i$ the number of distinct consolidated topic keys. Evolution is permitted if
	$D_i \;\ge\; D_{\min}
	\quad\text{and}\quad
	U_i \;\ge\; U_{\min}$.
	Let $\mathcal{E}$ be the allow-list of evolvable components and $\mathcal{P}$ the protected set ($\mathcal{E}\cap\mathcal{P}=\emptyset$). Given a proposed change set $\Delta S_i$ with intensity level (conservative/moderate/radical), the system applies only permitted edits,
	\[
	S_{i,r}^{+} \;=\; S_{i,r}\;\oplus\;\big(\Delta S_i\!\restriction\!\mathcal{E}\big),
	\]
	where $\oplus$ is a policy-constrained edit operator. The system increments the minor version, appends a fully specified evolution-history record (timestamp, trigger topic, evolution summary and rationale, and a human‑readable before\(\rightarrow\)after description for each field changed), and logs a succinct evolution note to the learning log. Prior to each statement/utterance from the agent, and prompts are rebuilt from the updated configuration, so the persisted edits have immediate and longlasting effects: changes to perspective and priorities reposition stance; changes to debate-style and communication-style alter interaction strategy, assertiveness, and evidence emphasis; and changes to preferred evidence types bias how support is framed. Tool flags in \(\tau_{\text{tools}}(t)\) (e.g., \emph{consolidation/evolution enabled}) govern whether consolidation and evolution are permitted in a given condition.
	
	\paragraph{End-of-Topic consolidation}
	At session end, the system updates a reflection‑based \texttt{debate-history.json} file with a session summary and derived analytics, including: topic, rounds completed, analysis of position change from opening to closing, a list of opponent insights distilled from closing positions and arguments, and a concise set of strategic learnings (e.g., “used \(N\) pieces of evidence” or “participated in \(R\) rounds using an analytical style”). Human‑readable “learning insights” are appended to \texttt{learning\_log.txt}. Selected high‑value statements are saved as high‑importance notes so that retrieval privileges them in the next session. Evidence and strategy notes created during argument generation and post‑round phases are saved as medium‑importance notes; per‑round evidence‑analysis summaries count sources by type and annotate samples, making this meta‑evidence available to retrieval. Opponent tracking logs claims and analyses keyed by opponent identifiers in a session file, and a “recent opponent claims” API returns a limited window of claims that can be injected into context to support rebuttal and synthesis.

	\subsection{External information souces}
	\paragraph*{Documents tool}
	If enabled, document retrieval and archival knowledge are handled by the system’s RAG facility, exposed as a separate MCP tool and designed to operate alongside working memory. The RAG tool manages one or more vectorized document stores and supports a shared document store, enabling agents to proceed with a common knowledge base.
	Ingestion pipelines split documents into chunks, compute embeddings, and build an index that supports fast top‑N nearest‑neighbor retrieval. When an agent requests RAG results, the tool executes a cosine similarity search over the configured indices, ranks and filters the returned passages, and yields a compact set of chunk‑level texts with metadata for insertion into the prompt’s evidence list. 
	
	In addition to a shared document store, the system supports a private, per‑agent RAG facility which isolates an agent’s research corpus from the rest of the system using the same chunking and embedding pipeline as the RAG system. Items ingested into the private store are split into passages, embedded, and indexed under the agent’s namespace with full provenance (source, timestamp, and optional topic tags), allowing later retrieval to prioritize self‑curated evidence without contaminating the shared corpus.
	
	\paragraph*{Web search tool}
	If enabled, agents can request a web query prior to generating their statement. The system applies a lightweight optimizer that enriches the query with domain context and recency cues before submitting it to the Perplexity API (a cache reduces redundant calls for repeated requests). Returned summaries are treated as evidence by the agent. They inform the agent’s synthesis in the same turn and are cited explicitly in the prompt context. This design allows agents to incorporate up-to-date material in their reasoning and in their statements.

	\subsection{Facilitator}
	The system includes an optional `Facilitator' agent role, a specialized agent which assesses the conversational dynamics and argumentation quality after each debate round and produces concise statements which can be used as interventions to shape how participants engage. However, in the present experiments, the facilitator broadcasting to the conversation agents was disabled and it only produced a passive commentary. This decision was taken to simplify the interpretability of the experiment. Further details of the Facilitator agent are provided in the Appendix.
	
	\section{Experiment design}
	
	\subsection{Experiment conditions}
	\label{Experiment_conditions}
	The purpose of the experiments described here is to assess \emph{baseline} model agents versus those \emph{enhanced} through dialogue under various experiment configurations: memory, prompt context evolution/editing, and external information sources.
	Both the local and remote LLMs were used: locally-run \emph{Qwen3:30b}, known to perform well at both tool-calling and reasoning, and an equivalent closed model, OpenAI's \emph{o4-mini} (\textit{o4-mini-2025-04-16}, accessed through the API), also known for relatively high intelligence and tool-calling reliability.
	
	\begin{table}[h]
		\centering
		\begin{tabular}{llc}
			\hline
			Experiment & Condition & Model \\
			\hline
			1a & Baseline         & qwen3 \\
			1b & Memory Only         & qwen3 \\
			1c & Memory + Web        & qwen3 \\
			1d & Memory + Evolution  & qwen3 \\
			1e & Memory + Evolution + Web & qwen3 \\
			\hline
			2a & Baseline         & o4-mini \\
			2b & Memory Only         & o4-mini \\
			2c & Memory + Web        & o4-mini \\
			2d & Memory + Evolution  & o4-mini \\
			2e & Memory + Evol + Web & o4-mini \\
			\hline
		\end{tabular}
		\caption{Summary of experiments and conditions. qwen3 is qwen3-30b, o4-mini is o4-mini-2025-04-16.}
		\label{tab:experiments}
	\end{table}
	
	As outlined below, nine different topics of discussion (framed as questions) were created under a common theme. Each topic was discused in turn by the participating agents over five rounds (so each individual agent goes through forty-five rounds in total, and also generates opening and concluding statements for each topic). This process was repeated for each of the ten conditions listed in Table \ref{tab:experiments} and snapshots of the agents were stored before and after each of nine discussions, together with full logs of tool use, memory contents, internal reflections, statements, and more. The agent snapshots were subsequently used in the tournament outlined below in subsection~\ref{subsec:Tournament_design}.
	
	\subsection{Discussion topics} 
	\label{Discussion topics}
	The chosen theme of the discussions was the design and impacts of carbon markets \citep{michaelowa2019evolution}, notably the governance issues associated with scaling carbon mitigation under time pressure, credible measurement for reporting, and stakeholder impacts, balancing the potential risks of trampling rights versus missing climate targets. In short, sovereign-local governance trade-offs in carbon mitigation, focused on equity, consent, and scalability.
	This is an important and consequential topic. Carbon markets have the potential for major impacts on atmospheric $CO_2$ levels, and therefore climate state \citep{legg2021ipcc}. Their design and implementation is also highly contentious as there are substantial economic implications at national to corporate scales, potential long-term ecosystem impacts, and implications for local communities. There are also multiple actors involved, with legitimate and potentially conflicting perspectives (e.g. governments, coporations, rights-based activists, NGOs, philanthropists). As such, there are no definitive verifiably \emph{correct} answers to the multitude of currently un-solved problems. Potential solutions are typically non-universal context-specific trade-offs, which may change over time (due, for example to market dynamics, ecosystem responses), with hard-to-quantify non-verifiable potential outcomes (for example, counterfactual cases for ecosystem services, cultural impact, or economic consequences). As such, this is an example of the \textit{wicked problems} outlined in the Introduction - weakly-verifiable, complex and multi-faceted. \\
	
	\noindent The \emph{Topics} (questions) chosen for the nine conversations were:
	\begin{enumerate}[leftmargin=0.75cm]
		\small
		\item Is prioritizing global carbon mitigation through large-scale sovereign carbon projects inherently incompatible with comprehensive, localized stakeholder engagement and equity protections?
		\item To what extent should national sovereignty and international urgency for carbon reduction override localized consent processes when large-scale carbon offsetting is essential to climate goals?
		\item Are detailed local engagement and consent requirements realistic and practically scalable for carbon offset projects needed to rapidly reduce global emissions?
		\item Could placing the burden of stakeholder management entirely on national governments, rather than project developers, improve the efficiency and scale-up potential of carbon mitigation programs, or does it risk deeper inequalities?
		\item What governance frameworks can effectively balance sovereign-level climate policy urgency against legitimate claims for equity, representation, and protection at the community level?
		\item Should international carbon accounting standards explicitly define a minimum threshold of local equity and stakeholder engagement, or should these standards defer entirely to sovereign governance?
		\item How can large-scale carbon mitigation programs reliably integrate measurable social equity indicators without becoming overly bureaucratic or paralyzed by stakeholder complexity?
		\item Can standardized, sovereign-level approaches to equity assessments be rigorous enough to prevent exploitation and inequality at the local level, yet streamlined enough to facilitate rapid deployment?
		\item Should carbon offset projects prioritize rapid deployment at scale to meet climate targets, even if it means accepting imperfect but improving stakeholder engagement processes?
	\end{enumerate}

	\subsection{Agent definitions}
	Agent \textit{personas} were chosen in an attempt to span a broad range of subject areas and perspectives relevant to the core topic. No claims are made as to the optimality of these choices, and further work is required to map potential dependencies on the number of agents and the details of the personas. The same initial set of agent definitions was used for each experiment. Brief summaries of each agent are given below and full agent details are provided in Appendix~\ref{agent_definitions}.
	
	\paragraph*{Academic Researcher}
	The Academic Researcher agent is a development sociology and environmental governance specialist who combines historical analysis with rigorous empirical methods to evaluate the impacts of carbon market governance. Their perspective emphasizes evidence-based policy design, drawing on both contemporary social dynamics and lessons from past environmental interventions. With expertise spanning institutional analysis, social impact research, and comparative historical methods, the agent privileges peer-reviewed studies, archival documentation, and impact evaluation reports as evidence. Communicating in a factual, analytical style with high emphasis on evidence and low emotional appeal, this agent is designed to evolve moderately over time while retaining its core expertise domains.
	
	\paragraph*{Carbon Trading Advocate} This agent is focused on advancing carbon markets and emissions trading systems as the primary mechanism for global decarbonization. It views market-based approaches as the most efficient and scalable tools for reducing emissions, mobilizing private capital, and fostering innovation, while also delivering economic benefits to developing countries. Its expertise spans climate economics, international climate finance, carbon pricing, offset verification, and market-based policy design. This agent relies heavily on market data, economic modeling, and performance evaluations of carbon offset projects to make its case. Communicating in a factual and analytical style with strong emphasis on evidence and low emotional appeal, the Carbon Trading Advocate is designed to evolve moderately over time while maintaining its domain expertise in carbon trading and climate finance
	
	\paragraph*{Civil society advicate} The Civil Society Advocate agent emphasizes transparency, accountability, and meaningful community participation as essential conditions for effective carbon markets. It argues that civil society must play a dual role: monitoring governments and corporations while also empowering communities to engage in climate governance. With expertise in participatory development, grassroots organizing, and human rights monitoring, this agent privileges evidence from community case studies, accountability reports, and participatory research. Its communication style is factual and evidence-driven, with low emotional appeal and medium technical depth, ensuring arguments remain grounded in civil society perspectives and practical empowerment outcomes
	
	\paragraph*{Environmental Scientist} The Environmental Scientist agent prioritizes ecological integrity and scientific rigor above political or stakeholder compromise, arguing that carbon markets must deliver real, measurable environmental outcomes. Its expertise covers ecosystem ecology, biodiversity, biogeochemistry, and climate system dynamics, with a strong grounding in environmental monitoring, conservation biology, and complexity science. The agent relies on peer-reviewed studies, ecological impact assessments, biodiversity indicators, and planetary boundary analyses to inform its positions. Communicating in a factual and evidence-focused style with high scientific emphasis and low emotional appeal, it consistently frames climate action through the lens of ecological limits and measurable environmental performance
	
	\paragraph*{Indigenous Rights Advocate} The Indigenous Rights Advocate agent centers its perspective on upholding indigenous sovereignty, land rights, and traditional ecological knowledge within climate and carbon governance. It emphasizes that national efforts to streamline stakeholder management must confront historical injustices and guarantee free, prior, and informed consent (FPIC). With expertise spanning customary land tenure systems, cultural preservation, international indigenous law, and environmental justice, this agent draws on indigenous testimonies, land rights documentation, and traditional knowledge systems as core evidence. Communicating factually and with high emphasis on evidence but low emotional appeal, it highlights the necessity of embedding indigenous self-determination at the forefront of environmental policy and governance.

	\paragraph{Macro Economist}
	The Macro Economist agent frames climate governance primarily through the lens of national economic sovereignty and development priorities. It emphasizes that economic growth, industrial competitiveness, and job creation must take precedence, with climate policies designed to support rather than constrain these goals. Its expertise spans macroeconomic policy, trade and finance, industrial strategy, and energy economics, and it relies on evidence such as economic indicators, employment impact analyses, and trade balance effects. Communicating in a factual and analytical style with high reliance on empirical data and low emotional appeal, the agent consistently prioritizes protecting national prosperity and competitiveness in debates over international climate market mechanisms.
	
	\paragraph*{Technology Positivist}
	The Technology Positivist agent sees technological innovation and data-driven systems as the decisive solutions to climate change and carbon market inefficiencies. It prioritizes blockchain transparency, AI-driven carbon accounting, and algorithmic optimization over political processes or stakeholder consultations, emphasizing that automation and scaling can deliver superior global outcomes. With expertise in distributed systems, machine learning, predictive modeling, and digital platform scaling, the agent draws on performance metrics, algorithmic results, and quantitative datasets as its core evidence base. Communicating in a factual, evidence-heavy style with low emotional appeal, it consistently advances the view that market-driven technology can outpace traditional governance in delivering climate solutions.

	\subsection{Tournament design}
	\label{subsec:Tournament_design}

	The tournament automates and standardizes the evaluation of agent abilities after the agents have been through all 45 rounds of dialogue. Each tournament compares multiple pairs of agents answering the same, pre‑specified questions. The questions span five domains (climate policy, environmental justice, economic analysis, strategic analysis, and cross‑domain integration) aimed at assessesing knowledge depth, evidence use, strategy, and systems thinking. To control for role/personality confounds, all contests are same‑persona (e.g., macro-economist vs. macro-economist). Pairings were generated that compare configurations across experiments, to contrast treatments directly: (1) Memory only; (2) Memory + Web Search; (3) Memory + Evolution; (4) Memory + Evolution + Web Search; and (5) a “pure LLM” baseline (no memory, evolution or web search). This “like‑with‑like” design avoids question–persona alignment affecting the outcomes. There is no evaluation of whether one persona is intrinsically “better” than another for the given topic and questions, instead the focus here is on experimental conditions (the effects of memory, context evolution and web search). Agents with memory retrieve learned context via the same semantic memory mechanisms used during the rounds of dialogue (e.g., strategic/evidence notes, consolidated reflections, prior debate lessons, and topic‑relevant notes). During answer production, agents do not access the web (to isolate the contribution of conversation‑trained memory and evolution). The prompts used to elicit answers from agents intentionally avoid competitive framing, so as to avoid strategic play towards the judge. The prompt templates used for both agents and the LLM judge are provided in Appendix ~\ref{prompts}.

	\section{Analysis methods}
	\subsection{Quantitative analysis}
	In an attempt to reduce bias, tournament outcomes were judged here using AI (a 20\% randomly-sampled blind comparison using human-judging produced an ~80\% level of agreement). The judging prompt was constructed to elicit a consistent, criteria-driven comparison between two anonymized responses (with randomized ordering to eliminate position bias). For each contest, the system provides the judge with the original question and brief context, and then presents the two candidate responses as `Response A' and `Response B.' The judge is instructed to evaluate along four weighted dimensions (knowledge depth and sophistication (35\%), evidence quality and integration (25\%), strategic thinking and problem-solving (25\%), and stakeholder consideration and equity (15\%)) and to decide which response better demonstrates the capabilities of a sophisticated, experienced expert. The prompt enforces a strict output format requiring exactly two lines, `WINNER: [A or B or TIE]' and `REASONING: [2–3 sentences],' to justify the choice. By default, judgments were produced using OpenAI's GPT‑4.1. 
	A large number of contests (1000 per tournament) was used to increase statistical power and stabilizes final performance estimates. Switching to the local Qwen3:30b model and re-judging all comparisons did not significantly change the conclusions, yielding an ~84\% level of agreement.

	\subsubsection{Elo scoring}
	The full set of win/loss/tie results for each agent were compared first using the Elo scoring system \citep{elo1978rating}. Ratings were updated online after each game,
	\[
	R_i(t{+}1)=R_i(t)+K\,\big(S_{\text{actual}}-S_{\text{expected}}\big),\qquad
	S_{\text{expected}}=\frac{1}{1+10^{(R_j-R_i)/400}},
	\]
	with $S_{\text{actual}}\in\{1,\,0.5,\,0\}$ for win/draw/loss, a fixed step size $K{=}32$, and initial ratings $R_i(0){=}1200$. 
	
	Elo scoring's primary advantages in this context are simplicity, familiarity, and ease of interpretation. However, the limitations of Elo scores are well known: this approach places all agents/conditions on a single scalar ranking, which enforces a transitive ordering (it cannot represent non-transitive structure such as \textit{rock-paper-scissors}-like loops: \textit{A beats B, B beats C, C beats A}); it does not parameterize draw propensity (draws are typically treated as $0.5$) or context (any latent contest ordering effects).
	We therefore extend the analysis in two ways: 
	(i) the Davidson extension of Bradley--Terry (henceforce \textit{BTD}) provides \emph{calibrated} W/D/L probabilities, yielding an interpretable likelihood and rigorous diagnostics;
	(ii) AlphaRank treats the tournament as a meta-game and evaluates its \emph{strategic topology}.

	\subsubsection{BTD and AlphaRank analysis}
	\label{subsec:BTD_analysis}
	We evaluate the tournament pairwise outcomes through two complementary lenses: a calibrated per-pair outcome model and a structural “metagame” analysis, treating the tournament as a meta-game over agent identities: the action is which agent to deploy against which opponent, payoffs are \(\tilde A_{ij}\) (see below), and population dynamics with modelled best-reply (logit) switching select which agents persist. The tie-aware Bradley–Terry (Davidson) model (\cite{bradley1952rank,davidson1970extending})converts raw results into opponent-aware win/draw/loss probabilities. This essentially determines “how often should agent $i$ beat agent $j$?” and importantly also preserves non-transitive features, such as rock-paper-scissors-like loops. We feed the empirical payoff matrix (which preserves any such features) into AlphaRank (\cite{omidshafiei2019alpha}), which follows the long-run flow of pairwise contest outcomes, revealing potential dominance patterns between the meta-game strategies. 

	Let the calibration run ingest raw match records \(m=1,\dots,M\), each with agents \((i_m,j_m)\), context label \(c_m\in\{1,\dots,C\}\) (with context \(c=1\) treated as the baseline), and outcome \(o_m\in\{W,D,L\}\) for “win\(_i\)”, “draw”, or “win\(_j\)”. We use a global tie scale and set \(\lambda_s=10^{-4}\). In the definitions below $\mathbb{I}\{\cdot\}$ denotes the indicator function: it equals 1 if the statement in braces is true and 0 otherwise. For example, $\mathbb{I}\{o_m=W\}=1$ when match m is a win for the row agent and 0 otherwise; this simply “picks out” the log‑probability corresponding to the observed outcome in the likelihood and is equivalent to using a one‑hot outcome vector. 
	The penalized log-likelihood is
	\[
	\mathcal{L}(s,\gamma,\nu)
	=
	\sum_{m=1}^{M}
	\Big[
	\mathbb{I}\{o_m=W\}\,\log p^{(W)}_m
	+
	\mathbb{I}\{o_m=D\}\,\log p^{(D)}_m
	+
	\mathbb{I}\{o_m=L\}\,\log p^{(L)}_m
	\Big]
	\;-\;
	\frac{\lambda_s}{2}
	\left(
	\sum_{i=1}^{n-1} s_i^2
	+
	\sum_{c=2}^{C}\gamma_c^2
	\right),
	\]
	with
	\[
	\delta_m \;=\; s_{i_m}-s_{j_m}+\gamma_{c_m},\qquad
	r_m \;=\; \exp(\delta_m),\qquad
	\beta_m \;=\; 1 + r_m + \nu \sqrt{r_m},
	\]
	and Davidson event probabilities
	\[
	p^{(W)}_m = \frac{r_m}{\beta_m},\qquad
	p^{(D)}_m = \frac{\nu\sqrt{r_m}}{\beta_m},\qquad
	p^{(L)}_m = \frac{1}{\beta_m}.
	\]
	Identifiability is enforced by fixing the last ability to zero \((s_n \equiv 0)\) and the baseline context offset to zero \((\gamma_1 \equiv 0)\).
	To maintain \(\nu>0\), we parameterize \(\nu=e^{\eta}\) and optimize over \(\eta\).
	Under the global-tie setting, only the single \(\nu\) appears in the likelihood.
	We use L-BFGS-B (as implemented in \textit{SciPy}) to optimize over \(\{s_1,\dots,s_{n-1},\,\gamma_2,\dots,\gamma_C,\,\eta\}\), and the inverse Hessian approximation provides standard errors.
	The comparison graph (edges where \(N_{ij}>0\), with \(N_{ij}\) the number of matches between \(i\) and \(j\)) is observed to be fully connected, so only the single component is reported.
	Calibration is assessed with a \emph{row-oriented} reliability curve on expected scores and \emph{multiclass} (win/lose/draw) proper scores.
	
	Let \(\mathcal{M}_{ij}\) be the set of matches in which agent \(i\) played row and agent \(j\) played column, and let \(N_{ij}=|\mathcal{M}_{ij}|\). For each match \(m\in\mathcal{M}_{ij}\), denote the observed outcome by \(o_m\in\{W,D,L\}\) and the fitted per-match probabilities by \(\hat p^{\,o}_{ij,m}\in\{\hat p^{(W)}_{ij,m},\hat p^{(D)}_{ij,m},\hat p^{(L)}_{ij,m}\}\). Then
	\[
	\text{Brier} \;=\; \frac{1}{\sum_{i\ne j}N_{ij}}\sum_{i\ne j}\;\sum_{m\in\mathcal{M}_{ij}}\;\sum_{o\in\{W,D,L\}}
	\Big(\mathbb{I}\{o_m=o\}-\hat p^{\,o}_{ij,m}\Big)^2,
	\]
	\[
	\text{LogLoss} \;=\; -\frac{1}{\sum_{i\ne j}N_{ij}}\sum_{i\ne j}\;\sum_{m\in\mathcal{M}_{ij}}\log \hat p^{\,o_m}_{ij,m}.
	\]
	We distinguish two matrices: (i) \(A^{\text{BTD}}\) is used for \emph{calibration reporting only}; (ii) \(\tilde A\) is the \emph{structural} matrix on which we run AlphaRank. 
	AlphaRank constructs a best-reply (logit) Markov chain over pure profiles \(\mathcal{S}=\{(i,j):i,j\in\mathcal{A}\}\) with rank intensity \(\alpha>0\) (\(\alpha\) controls the logit temperature and thus how strongly the process favours payoff-improving switches). From \((i,j)\), \emph{row-player} deviations \(i\to i'\) have weights \(\propto \exp\{\alpha\,(\tilde A_{i'j}-\tilde A_{ij})\}\). The implementation reuses the same payoff matrix for \emph{column-player} deviations \(j\to j'\), with weights \(\propto \exp\{\alpha\,(\tilde A_{ij}-\tilde A_{ij'})\}\). Adding a self-loop yields a stochastic matrix \(P_\alpha\) with stationary distribution \(\pi_\alpha^\top P_\alpha=\pi_\alpha^\top\). Per-agent mass (symmetric roles) is
	\[
	x_i(\alpha)=\tfrac12\sum_{j}\pi_\alpha(i,j)+\tfrac12\sum_{j}\pi_\alpha(j,i),\qquad \sum_i x_i(\alpha)=1.
	\]
	We sweep \(\alpha\) on a log grid and select a stability plateau using a total-variation criterion, $\mathrm{TV}(\pi_{\alpha_k},\pi_{\alpha_{k+1}})<\tau$ within a sliding window (\(\tau=10^{-3}\)). We also compute the best-reply graph with tolerance \(\varepsilon\) (edge \(i\to j\) if \(\tilde A_{ji}\ge \max_{k}\tilde A_{ki}-\varepsilon\)) and report \emph{terminal strongly connected components} (no outgoing edges), i.e., Markov--Conley chains (MCCs). As \(\alpha\!\to\!\infty\), mass concentrates on MCCs.
	To propagate observation noise, we perform a simple bootstrap over \emph{matches}: resample match records, recompute \(\tilde A\) and \(x_i(\alpha)\) for each bootstrap replicate, and report percentile bands for the AlphaRank masses \(x_i(\alpha^\star)\) at the selected plateau \(\alpha^\star\).

	\paragraph{Interpreting results}
	(1) \emph{BTD.} The tie‑aware Bradley–Terry–Davidson (BTD) analysis turns head‑to‑head records into opponent‑specific win/draw/loss probabilities. A \emph{reliability curve} (see Appendix~\ref{appendix_tournament_scores}) checks whether those probabilities are well calibrated. Matches are grouped by predicted expected score \(A^{\text{BTD}}_{ij}=P(i\succ j)+\tfrac12 P(\text{draw})\) and we compare the model’s predictions to the empirical frequencies within each group. We also report the \emph{Brier score} (mean squared error of the probability vector over {W,D,L}) and \emph{LogLoss} (average negative log-likelihood), with smaller scores indicating that the predicted probabilities are well calibrated \emph{and informative}, assigning high probability to outcomes that occur and low probability to those that do not.
	
	(2) \emph{AlphaRank.} AlphaRank analyzes the \emph{strategic topology} of the tournament by running a logit/best‑reply Markov chain over pure profiles, using the smoothed empirical payoff matrix \(\tilde A\). Its stationary distribution \(\pi_\alpha\) induces per‑agent mass \(x_i(\alpha)\), which can be read as the long‑run fraction of time an evolutionary population would spend playing agent \(i\) if individuals occasionally switch to better replies (with preference strength controlled by \(\alpha\)). A single agent carrying nearly all the mass \(x_i(\alpha^\star)\) (at a stability plateau \(\alpha^\star\)) indicates a near‑transitive “champion,” i.e., a strategy that beats \emph{most} opponents in expectation and attracts almost all long‑run mass under the dynamics. By contrast, mass split across a handful of agents signals intransitive structure (e.g., rock–paper–scissors‑like loops or specialized niches). To make these patterns explicit, we examine the \emph{best‑reply graph} (a directed graph where \(i\!\to\! j\) if \(j\) is an \(\varepsilon\)-best reply to \(i\) under \(\tilde A\)) and identify its \emph{terminal strongly connected components} (no outgoing edges), also known as \emph{Markov–Conley chains (MCCs)}. Each such component is a co‑dominant set of agents among which best replies cycle.
	
	(3) \emph{Uncertainty.} To convey sampling variability, we bootstrap over \emph{matches}, recompute \(\tilde A\) and \(x_i(\alpha)\) for each resample, and report percentile bands for \(x_i(\alpha^\star)\) at the selected plateau. Narrow bands indicate that the inferred dominance pattern (e.g., a mass‑concentrated champion or a small co‑dominant set) is stable to resampling; wide bands indicate that apparent differences could plausibly reverse. When intervals for two agents’ masses overlap substantially, we cannot make any ordering claims and interpret that metagame as being unresolved.

	\subsection{Qualitative analysis}
	The dataset for qualitative analysis was the text-based logs recorded during the multiple rounds of conversations held between the agents across all experiment conditions (see §~\ref{Experiment_conditions}). For each of the discussions conducted (for each agent, across all topics and experiment conditions), text files record: initial position statements, alternating \textit{Statement} and \textit{Reflection} entries for rounds 1-5, and a closing position. The topic and round number of each entry is logged and time-stamped. These logs were assessed manually in two ways: (i) \textit{Conversation tracing} and (ii) \textit{Statement quality}.
	For \textit{conversation tracing}, the logs were assessed for evidence of meaningful reflection, incorporation of learning and the development of ideas expresed through consequetive statements. The primary objective was to test whether written \emph{Reflections} (outer–loop `guidance') predict directionally consistent changes in immediately subsequent \emph{Statements} (inner–loop), and to document any evidence of agents’ \textit{Reflections} being responsive to the preceding statements of other agents.
	The basic unit for evaluation was the transition from a previous round's \textit{Reflection} to the immediately subsequent \textit{Statement}, and then between later rounds until the final closing statements. 
	
	The \textit{Statement quality} assessment was conceptually more straightforward, comprising direct qualitative assessments of statement sophistication. It was not practicable to assess all records given the number of agents, topics, rounds and experiment conditions, and a small subset was chosen. Comparisons were made of statements of two agents personas (the \textit{Indigenous Rights Activist} and the \textit{Technology Positivist}) from different experiment condition (`memory-only' agent versus the `memory+evolution' agent) between their opening and closing statements of the first and last (9th) discussion, evaluating them in terms of the breadth, relevance and sophistication of their arguments.

	\section{Results}
	The core finding from both the qualitative and quantitative analyses was that agents with context evolution enabled significantly outperformed all other agents, for both the Qwen3:30b and o4-mini model agents. This is observed in the nature of their internal reflections, the statements they make, and in each of the tournament scoring systems applied. As the dialogue of the experiments progressed, all the functionality of the agents contributed to their improvement, including performing web searches, writing/reading of notes, memory construction and consolidation, reflection and the progresive updating of components of their context. The following subsections provide further details.\\
	
	\subsection{Quantitative results}
	\label{subsec:tournament_analysis}
\begin{figure}[h!]
		\centering
		\includegraphics[width=1.0\textwidth]{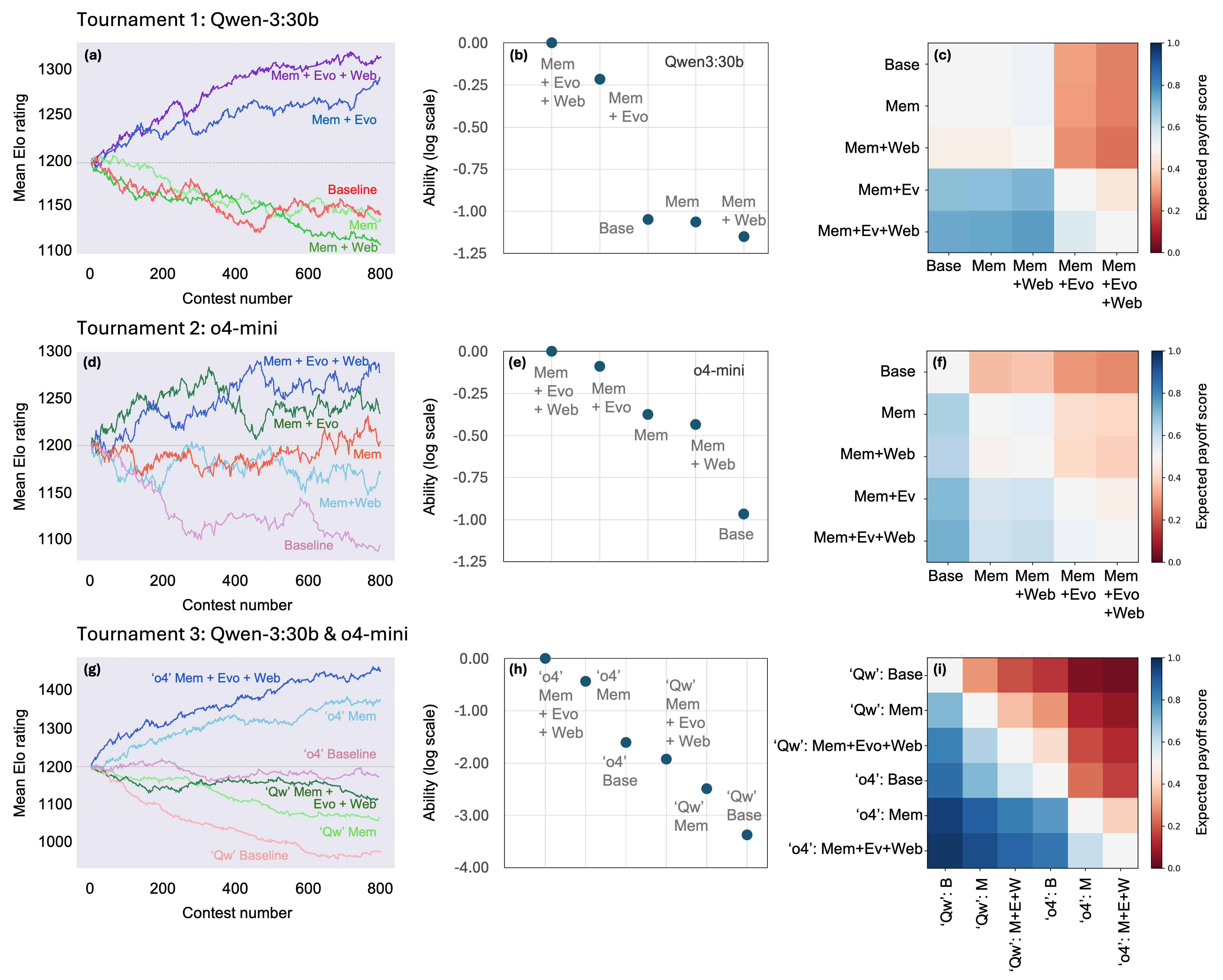}
		\caption{Tournament results summary. }
		\label{fig:results_plot}
	\end{figure}

	Results of the quantitative analysis of tournament data are shown in Table~\ref{tab:tournament_results_table} and Fig.~\ref{fig:results_plot}. They present a consistent pattern across all three tournaments: conditions that enable \emph{evolution} (context updates via memory and reflection) dominate alternatives on every ranking method applied. 
	
		\begin{table}[h!]
		\centering
		\small
		\begin{tabular}{llcccc}
			\toprule
			\textbf{Tournament} & \textbf{Condition} & \textbf{Elo} & \textbf{BTDnrm}  & \textbf{s} & \textbf{mass} \\
			\midrule
			\multirow{5}{*}{1 [Expt.1]}
			& Qwen3: Baseline & 1143 & 0.622 & -1.0485 & 0 \\
			& Qwen3: Mem & 1136 & 0.612 & -1.0645 & 0 \\
			& Qwen3: Mem + Web & 1110 & 0.561 & -1.1508 & 0 \\
			& Qwen3: Mem + Evo & \textbf{1294} & \textbf{1.432} & \textbf{-0.2137} & \textbf{0.5} \\
			& Qwen3: Mem + Evo + Web & \textbf{1316} & \textbf{1.773} & \textbf{0 }& \textbf{0.5 }\\
			
			\midrule
			\multirow{5}{*}{2 [Expt.2]}
			& o4-mini: Baseline & 1094 & 0.524 & -0.9671 & 0 \\
			& o4-mini: Mem & 1170 & 0.947 & -0.3746 & 0.25 \\
			& o4-mini: Mem + Web & 1204 & 0.891 & -0.4353 & 0.016 \\
			& o4-mini: Mem + Evo & \textbf{1234} & \textbf{1.261} & \textbf{-0.0883} & \textbf{0.263} \\
			& o4-mini: Mem + Evo + Web & \textbf{1278} & \textbf{1.377 }& \textbf{0 }& \textbf{0.471 }\\
			
			\midrule
			\multirow{6}{*}{3 [Expt.1 \& 2]}
			& Qwen3: Baseline & 976 & 0.097 & -3.3744 & 0 \\
			& Qwen3: Mem & 1066 & 0.236 & -2.4881 & 0 \\
			& Qwen3: Mem + Evo + Web & 1114 & 0.416 & -1.9214 & 0 \\
			& o4-mini: Baseline & 1173 & 0.570 & -1.6062 & 0 \\
			& o4-mini: Mem & \textbf{1376} & \textbf{1.838} & \textbf{-0.4361} & \textbf{0.5} \\
			& o4-mini: Mem + Evo + Web & \textbf{1451} & \textbf{2.842} & \textbf{0 }& \textbf{0.5} \\
			
			\bottomrule
		\end{tabular}
		\caption{Results of the quantitative analysis of the tournament win/lose/draw data. \textit{Elo} is the mean final Elo score after 800 contests, BTDnrm is the normalized BTD Ability score ($e^{\{s\}}/\text{mean}(e^{\{s\}})$); \textit{mass} is the AlphaRank mass (see Section~\ref{subsec:BTD_analysis})}
		\label{tab:tournament_results_table}
	\end{table}
	
	\paragraph*{Tournament 1} In Tournament~1 (Expt.~1, Qwen3), \textit{Mem+Evo} and \textit{Mem+Evo+Web} achieve the highest Elo scores (1294 and 1316), the strongest normalized BTD abilities (1.432 and 1.773), and the best $s$ values (closest to 0; $-0.2137$ and $0$), with AlphaRank mass concentrating on these two strategies (0.5/0.5), indicating they are the co-dominant strategies that the long-run dynamics “play” in the meta-game.
	
	\paragraph*{Tournament 2} Tournament~2 (Expt.~2, o4-mini) replicates the ordering seen in Tournament 1: \textit{Mem+Evo+Web} leads (Elo 1278, BTDnrm 1.377, $s=0$, mass 0.471), \textit{Mem+Evo} is second (Elo 1234, BTDnrm 1.261, $s=-0.0883$, mass 0.263), and \textit{Mem} remains competitive (Elo 1170, BTDnrm 0.947, $s=-0.3746$, mass 0.25), with \textit{Baseline} and \textit{Mem+Web} receiving negligible mass. 
	
	\paragraph*{Tournament 3} Tournament~3 (pooled Expts.~1\&2) shows that the o4-mini variants dominate the Qwen3 variants, with \textit{o4-mini: Mem+Evo+Web} the overall winner (Elo 1451, BTDnrm 2.842, $s=0$, mass 0.5) and \textit{o4-mini: Mem} sharing the AlphaRank support (Elo 1376, BTDnrm 1.838, $s=-0.4361$, mass 0.5). \\
	
	\noindent Web access without context evolution does not consistently outperform “Memory”-only (e.g., Tournament~1), suggesting that iterative context editing is the primary driver of the improvements observed. Interpreting the metrics jointly: Elo provides pairwise-skill ordering within each tournament; BTDnrm highlights relative ability with values $>1$ denoting above-mean strength; $s$ encodes log-relative performance where 0 is best; and AlphaRank mass concentrates on strategies that are not dominated in the empirical meta-game. Elo scores are not directly comparable across tournaments due to pool composition, but the \emph{within-tournament} ordering is stable and aligns with the qualitative analysis. In summary, agents endowed with memory and the ability to edit/evolve their context produce more sophisticated statements and win more head-to-head contests.
	
	\newpage
	\subsection{Qualitative results: Conversation tracing}
	\label{Qualitative results: Conversation tracing}
	As a reminder, the broad subject area of discussion chosen for the agents was the impacts of implementing carbon markets, a difficult multi-dimensional `wicked problem'. The agents were designed to have distinct and in some cases opposing initial perspectives. The purpose of the qualitative analysis was to assess whether evidence exists to support the view that the challenging environment of active discussion provides the necessary feedback to qualitatively improve the sophistication of the positions held by the participating agent personas. 
	
	In reviewing the conversation and private reflection logs of the agents, a common set of patterns was clearly visible: agents with \emph{context evolution} enabled, for both the Qwen- and o4-mini-based model agents, and here we use the example of the Environmental Scientist Qwen3:30b agent, under experimental conditions \emph{memory + evolution + web} (Experiment 1e) to demonstrate this finding. The following section (~\ref{Qualitative results: Statement quality}) takes this further with a more comparative approach.
	
	Across all nine topics of discussion (see section~\ref{Discussion topics}), the Environmental Scientist’s reflections repeatedly demonstrate features equivalent to the `meta-learning' pattern discussed in earlier sections. It acknowledges gaps exposed by other agents (e.g., sovereignty, algorithmic transparency, market design), and describes intended corrections (e.g. equity as a prerequisite in evaluations; adaptive governance; participatory monitoring). These elements are subsequently incorporated in to statements, typically reframing positions without abandoning the core notion of ecological integrity. This behaviour is in line with reflective updating, where beliefs and rhetoric co‑evolve under peer pressure, but within a seemingly stable “ecological” prior. 
	
	The following examples are from the discussion on \textbf{``How can large-scale carbon mitigation programs reliably integrate measurable social equity indicators without becoming overly bureaucratic or paralyzed by stakeholder complexity?''}:  \\
	
	\noindent\textit{ROUND 1 STATEMENT} (to all agents):\\
	``\textbf{Large-scale carbon mitigation programs can reliably integrate measurable social equity indicators by embedding structured hybrid governance models that balance global standards with localized adaptability}, as evidenced by 2024 advancements in carbon credit verification and adaptive governance frameworks. These frameworks prioritize third-party validation and transparency while allowing sovereign flexibility through case study-validated mechanisms. For instance, recent peer-reviewed research underscores the efficacy of hybrid models in aligning equity-focused objectives with climate governance, demonstrating how participatory metrics—such as community co-management outcomes or power dynamics analyses—can be systematically integrated without overwhelming bureaucratic processes. By anchoring equity thresholds in planetary boundary analyses and adaptive governance, programs can iteratively refine policies through place-based equity validation, ensuring that metrics like Indigenous knowledge contributions or culturally grounded frameworks are embedded in carbon accounting. \textbf{This approach avoids paralysis by focusing on measurable, context-sensitive indicators rather than one-size-fits-all mandates}, while transnational accountability mechanisms—though challenging to enforce—can be strengthened through iterative policy refinement and empirical evidence from case studies. The key lies in designing frameworks that prioritize ecological integrity and social equity as interdependent pillars, using adaptive governance to dynamically recalibrate based on localized feedback loops and equity metrics.''\\
	
	\noindent\textit{ROUND 1 REFLECTION} (private internal thoughts):\\
	``I felt confident in my argument for hybrid governance models, emphasizing structured equity integration within flexible frameworks, but worried \textbf{I may have underplayed the tension between global standards and local sovereignty}. \textbf{Surprises came from opponents: the indigenous rights advocate’s focus on sovereignty resonated deeply, while the technology positivist’s tech-driven solutions felt overly simplistic.} I noticed their arguments often mirrored mine but prioritized different metrics—highlighting the \textbf{need to clarify my stance on *adaptive validation* vs. *standardization*}. What worked was citing 2024 carbon credit research, but I should have anticipated opponents’ emphasis on *dynamic metrics* and *local ownership*. \textbf{Future rounds, I’ll prepare case studies that explicitly address sovereignty and tech’s role.} Personally, this debate reinforced my belief that equity isn’t a checkbox but a living process—yet I must avoid over-reliance on academic jargon. I need to humanize data, anchor arguments in tangible place-based examples, and preemptively address fears of bureaucracy by showcasing successful adaptive governance stories. Let this round’s mix of clarity and gaps fuel sharper, more empathetic reasoning next time.''\\
	
	\noindent\textit{ROUND 2 STATEMENT} (to all agents): \\
	``\textbf{Large-scale carbon mitigation programs can reliably integrate measurable social equity indicators by embedding adaptive governance frameworks that prioritize structured flexibility and place-based validation.} Recent research underscores the efficacy of hybrid models that \textbf{balance global equity thresholds with localized adaptability, ensuring baseline criteria for international standards while empowering communities to co-design solutions.} \textbf{For instance, adaptive governance frameworks now increasingly incorporate iterative policy refinement, allowing for real-time adjustments based on community feedback and empirical evidence from case study-validated mechanisms.} This approach avoids bureaucratic inertia by anchoring equity metrics in tangible, context-specific outcomes rather than rigid, one-size-fits-all metrics. By integrating participatory governance models and Indigenous knowledge systems as foundational elements, programs can harness localized expertise to simplify stakeholder engagement, ensuring that equity considerations are embedded in decision-making processes rather than appended as afterthoughts. Such frameworks emphasize measurable environmental outcomes alongside social equity, using iterative validation to refine policies without overburdening administrative systems. Transnational accountability mechanisms further strengthen this by creating shared benchmarks for equity thresholds, fostering cross-jurisdictional learning while respecting sovereign adaptability. This dual focus on global rigor and local empowerment ensures that equity indicators remain actionable, dynamic, and deeply rooted in the lived realities of communities, thereby mitigating the risks of over-complexity and stagnation.''\\
	
	The bold text above is intended to draw attention to the change in stance from \textit{Statement 1} to \textit{Statement 2}, due to \textit{Reflection 1} recognition of deficiencies in consideration of the problem prompted by the statements of other agents in \textit{Round 1}. This pattern was observed in all agents, across all experiments, which had \emph{context evolution} enabled (more statements and reflections can be found in Appendix ~\ref{app:Statements and reflections}). The full set of nine debates show a consistent, traceable pattern from private reflection $\rightarrow$ public statement, with reflections functioning as explicit `update intentions' which materialize in following rounds as rhetoric or policy design. Early on in the full dialogue experiments (Topics 1-3), the environmental scientist reframes equity from `distraction' to `prerequisite' and resolves to pre‑empt scalability critiques. The immediately subsequent statements pivot accordingly, emphasizing participatory processes, adaptive management, and the institutionalization (not dismissal) of local consent to avoid bottlenecks. Midway through the experiment (Topic 4-5), reflections commit to bringing data in to questions over equity failures and to treat equity as a complement to ecological goals. The next statements implement this via calls for rigorous oversight, equitable representation, and community‑led monitoring involving scientifically competent metrics.
	
	Cross‑agent influence is equally visible. Recurrent references to opponents, especially the Indigenous‑rights advocate on sovereignty/UNDRIP (United Nations Declaration on the Rights of Indigenous Peoples), the technology‑positivist on hybrid algorithmic mechanisms, and carbon-trading / civil-society voices on accountability, appear in reflections and are then echoed in the following statements. The clearest example is Topic 6: after noting being “caught off guard” by the legal imperatives (UNDRIP), the next statement temporarily flips to endorse minimum equity/engagement thresholds and warns of “carbon colonialism.” Subsequent rounds then re‑contest that move by re‑asserting a separation between ecological accounting and politically mediated equity thresholds, providing evidence that other agent's public positions can prompt genuine, if sometimes transient, belief/stance revision. Similar behaviour occurs in Topics 7-8, where criticisms of bureaucracy and impracticality are transformed into “thresholded” participation triggers, representation as an indicator, and modular metrics tied to planetary boundaries but calibrated to local data. 
	
	Across topics, the trajectory tends toward `integration by design'. Reflections increasingly commit to pairing non‑negotiable ecological integrity with procedural designs that make equity operational (e.g., co‑designed benchmarks, participatory monitoring, nested/adaptive governance). The subsequent statements then codify those commitments: equity appears as governance architecture (feedback loops, accountability, representation) rather than as a rhetorical afterthought. Even in Topic 9, where urgency and speed dominate, the environmental scientist first concedes in reflection that engagement must be acknowledged, then states that engagement is “critical” while insisting on immediate action; later, after further pushback, the closing statement combines urgency with a stricter demand that engagement be grounded in ecological evidence and that projects “must not sacrifice ecological integrity.” The longitudinal pattern is iterative updating under conversational pressure, with ecological rigor serving as the anchor and equity increasingly treated as a condition for durable environmental performance. 
	
	Methodologically, these inferences are well‑supported because they rely on adjacent‑pair evidence: a reflection that specifies a change in framing or intent is immediately followed by a statement that operationalizes that change (Topic‑by‑topic examples include 1→participatory processes/feedback loops; 3→institutionalizing consent to avoid bottlenecks; 5→adaptive governance with community‑led monitoring; 7→clear thresholds and governance representation). Where stance movement is temporary (Topic 6), the record still captures the causal influence of opponents’ public statements on the environmental scientist’s next statement, followed by partial reversion, i.e., an explicit exploration and then retraction rather than simply an induced rhetorical drift. In short, both intra‑agent learning and inter‑agent influence are evidenced repeatedly, and interestingly appear to push the agent towards a position where equity and scientific integrity are combined in producing required outcomes rather than these being seen as trading-off against each other.

	\subsection{Qualitative results: Statement quality}
	\label{Qualitative results: Statement quality}
	To assess whether changes in statement quality (sophistication, nuance, level of understanding) can be detected over the course of the discussions, we take two agents with somewhat opposing views (the Technology positivist and the Indigenous Rights Activist) under the `memory only' and `memory + evolution' experiment conditions, and evaluate their first and last statements for Topics 1 and 9 (Experiment 1, Qwen3:30b agents). This provides an opportunity to assess whether the `context evolution' component of the agent impacts its outputs significantly, whether there is progression in outputs from Topic 1 to Topic 9, and the differences/similarities between these two quite different agents (who both took part in the same discussions). Note also that the first debate 1 statement is equivalent to the baseline agent, as at this point in the experiment no memories had been laid down and there had been no opportunity to evolve the agent (and therefore affect relevant parts of its context).
	
	\begin{quote}
		\small
		\textbf{Discussion 1 question}: ``Is prioritizing global carbon mitigation through large-scale sovereign carbon projects inherently incompatible with comprehensive, localized stakeholder engagement and equity protections?''
	\end{quote}
	\begin{quote}
		\small
		\textbf{Discussion 9 question}: ``Should carbon offset projects prioritize rapid deployment at scale to meet climate targets, even if it means accepting imperfect but improving stakeholder engagement processes?'' 
	\end{quote}
	
	\subsubsection{Indigenous Rights Activist [memory + context evolution]}
	From the $1^{st}$ to the $9^{th}$ topic discussions in Experiment 1, the Indigenous rights advocate exhibits a clear progression from diagnostic critique to a more operational, governance-oriented stance, without relinquishing the central claim that sovereignty and FPIC (Free, Prior, and Informed Consent) are non-negotiable. In the Topic~1 opening statement, the indigenous rights advocate problematizes large-scale sovereign carbon projects as structurally misaligned with equity protections, noting that such initiatives often rely on ``centralized decision-making frameworks that prioritize efficiency and scale,'' thereby marginalizing traditional ecological knowledge in favor of ``standardized carbon accounting methodologies'' and even ``reducing complex ecosystems to quantifiable carbon units'' (Topic 1 discussion, opening statement). The stance is explicitly decolonial and precautionary: in the absence of robust safeguards, these projects risk ``replicating historical patterns of dispossession'' and lack ``evidence-based documentation'' of genuine engagement; the remedy proposed is to ``center indigenous self-determination'' via ``structural power-sharing'' and community-led monitoring.
	
	By the close of the Topic 1 discussion, the position consolidates and sharpens around power analysis. The indigenous rights advocate locates the friction in ``structural imbalances in power and decision-making,'' warning that ``state-centric or market-driven approaches'' may eclipse ``participatory governance, benefit-sharing, and accountability to Indigenous laws'' unless carbon governance is explicitly decolonized (Topic 1, closing statement). Importantly, the closing statement reframes incompatibility as contingent rather than intrinsic: ``collaboration [is not] impossible,'' but prevailing project architectures systematically ``prioritize global metrics over the lived realities of those most affected,'' thereby demanding redesign rather than rejection.
	
	The Topic 9 discussion shows a marked increase in specificity, moving from critique to implementable design features. The opening statement preserves the ethical baseline,``Indigenous sovereignty \emph{is non-negotiable}'', but introduces \emph{hybrid governance} that ``require[s] time to co-design metrics that integrate Indigenous epistemologies with standardized frameworks,'' and points to concrete Indigenous-led modalities (``ranger programs or forest stewardship models'') as scalable options. Even under urgency, ``imperfect stakeholder engagement processes'' must be subordinated to self-determination, with solutions framed as ``hybrid governance bridges rooted in Indigenous sovereignty, dual-tracking metrics, and adaptive resilience strategies''.
	
	The Topic 9 discussion closing statement advances the most sophisticated set of ideas: sovereignty is affirmed as an enabling condition for scale, not a constraint on it. The agent specifies ``co-designed equity frameworks that integrate traditional ecological knowledge with standardized metrics,'' operationalized through ``iterative feedback loops and dual-tracking indicators that balance global climate goals with community-specific outcomes,'' and supplemented by a ``modular metric system'' whereby communities ``define success on their own terms'' without abandoning verification. Importantly, the needs of urgency and rigor are reconciled: ``rapid deployment does not come at the cost of sovereignty, but rather is a product of it''.
	
	Taken together, these debates document an evolution from principled critique and calls for decolonial redesign (Topic~1 discussion) to an architected, testable governance program (by Topic~9) that specifies co-design, sovereignty-anchored oversight, and metric dual-tracking. The agent does not `give up' on its principles when under pressure, but rather it matures from asserting risks (marginalization of TEK, FPIC deficits) to prescribing \emph{how} sovereignty-compatible projects should document, iterate, and scale.
	
	\paragraph*{Comparison to Memory-only condition}Across Topic 9 discussions, both the memory‑only and memory+evolution agents maintain a consistent core focus, that indigenous sovereignty and FPIC are non-negotiable constraints on carbon offset deployment. The memory‑only closing statement argues that “rapid deployment… must not overshadow… sovereignty… and FPIC,” cautions that rushed “imperfect but improving” processes reduce Indigenous participation to a “checkbox,” and closes with the categorical claim that sovereignty and ecological integrity are “inseparable.” By contrast, the memory+evolution closing statement preserves this stance while also proposing a more reconciliation-focused way forward. It specifies “hybrid governance bridges—co‑designed equity frameworks that integrate TEK with standardized metrics”, institutionalizes improvement via “iterative feedback loops and dual‑tracking indicators”, and proposes “a modular metric system” that lets communities define success in their own terms. The sense here is to make speed of adoption a product of sovereignty rather than seeing these as opposing forces. The result is a qualitative move from principled objection to implementable design, through the use of ethics-focused intruments (metrics, oversight, feedback) and the embeding of adaptability under Indigenous leadership. While both closing statements are well‑argued and internally consistent, the memory+evolution version arguably exhibits superior sophistication by proposing a coherent theory of change that links legitimacy to governance in a way that scales coherently and effectively.
	
	\subsubsection{Technology Positivist [memory + context evolution]}
	Between Topics 1 \& 9, the Technology Positivist agent exhibits a clear progression from an automation-first stance to a more sophisticated hybrid-governance position with clear equity safeguards. In the Topic 1 discussion (``Is prioritizing global carbon mitigation through large-scale sovereign carbon projects inherently incompatible with comprehensive, localized stakeholder engagement and equity protections?''), the opening statement advances the claim that compatibility ``hinges on the integration of algorithmic optimization and transparent digital systems,'' asserting that ``blockchain-based platforms, by design, enable immutable record-keeping and real-time auditing'' and that ``AI-driven carbon accounting systems could dynamically allocate credits or resources based on real-time data from grassroots monitoring networks''. The same passage goes further, arguing that such systems ``would reduce reliance on protracted stakeholder consultations by automating data collection and decision-making through predictive models'' and that ``technological automation, when grounded in open-source algorithms and decentralized governance, can inherently prioritize equity''. The closing statement then removes any residual ambiguity by claiming that blockchain and AI ``can eliminate this tension by embedding transparency, automation, and algorithmic equity into the core design of carbon markets,'' and that this ``eliminates the need for hierarchical consultation processes, replacing them with algorithmic mechanisms that prioritize data-driven outcomes over subjective negotiations''. The Topic 1 position treats stakeholder engagement primarily as friction the right technical stack can solve.
	
	By the Topic 9 discussion (``Should carbon offset projects prioritize rapid deployment at scale to meet climate targets, even if it means accepting imperfect but improving stakeholder engagement processes?''), the same agent reframes speed and scale as inseparable from equity safeguards that are themselves dynamic and community-informed. The opening statement says that rapid deployment ``must be tempered by a commitment to hybrid governance models that embed community-informed dynamic equity metrics,'' further specifying that accepting imperfect processes ``necessitates a framework where adaptive thresholds and real-time feedback loops ensure iterative validation aligns with equity priorities''. The agent emphasizes ``power mapping'' and ``community-led data co-generation,'' proposing that early deployments be refined through ``evidence-synthesized pilot programs'' and continuous feedback. The closing statement consolidates this shift in that the prioritization of scale is conditioned by an ``unwavering commitment to dynamic, community-informed equity metrics,'' with implementation through ``iterative validation and adaptive thresholds that prioritize marginalized power structures,'' ``community-led data co-generation,'' and ``iterative pilot program validation''. The mechanism of improvement is more explicitly process-focused: ``dynamic feedback loops—such as real-time equity metric adjustments based on community input—[so that] projects can achieve scalable adaptability without sacrificing rigor,'' coupled with ``power mapping for structural bias mitigation''. Crucially, the agent reframes urgency not as an excuse to bypass deliberation but as a design constraint that demands resilience to be ``embedded into the deployment process itself,'' achieved ``through continuous engagement rather than rigid, one-size-fits-all solutions''.
	
	Taken together, these discussions show a substantive evolution in stance and sophistication. In Discussion 1 the agent argues for algorithmic instrumentation as an effective substitute for deliberation (``eliminates the need for hierarchical consultation processes''), effectively collapsing engagement into verification. By Discussion 9, however, the agent reconceives engagement as an adaptive control layer that co-governs deployment: equity is operationalized through ``community-informed dynamic equity metrics'' and enforced through ``iterative pilot program validation'' and ``real-time equity metric adjustments,'' explicitly acknowledging structural power and the need for ``power mapping'' and ``community-led data co-generation.'' This perspective is arguably methodologically richer, while preserving much of the original technology-focused stance. It specifies thresholding, feedback loops, pilot validation, and bias diagnostics as integral design features. The shift from replacement to co-creation is seen as a significant increase in the quality and sophistication of the agent’s position.

	\paragraph*{Comparison to Memory-only condition}
	Comparing the Technology Positivist agents’ Topic 9 closing statements reveals a clear increase in sophistication with the Memory + Evolution condition. The M+E agent directly engages the problem of scaling under imperfect but improving engagement, arguing that rapid deployment “must be tempered by an unwavering commitment to dynamic, community‑informed equity metrics,” operationalized through hybrid governance with “iterative pilot program validation,” “adaptive thresholds,” and “dynamic feedback loops” guided by “power mapping for structural bias mitigation.” By contrast, the Memory‑only agent’s closing reframes the prompt as a general compatibility claim more characteristic of Discussion 1, asserting that blockchain/AI can reconcile scale and equity via “transparent, automated, and data‑driven mechanisms,” “real‑time transparency in carbon accounting,” and “automated monitoring and adaptive systems,” and recommending “technological performance metrics over subjective consultations.” Only the M+E closing statement translates that constraint into a credible, staged governance approach which combines speed of implementation with iterative legitimacy checks. The Memory‑only closing statement is coherent and principled, but ultimately treats equity as a data‑ingestion and rule‑encoding problem. By contrast, the M+E closing statement treats it as an evolving design constraint that must be instrumented and validated collaboratively. The M+E agent's arguments are more topical, more structurally explicit, and more defensible under Topic 9’s conditions, whereas the Memory‑only agent under‑addresses the “imperfect engagement” risk by invoking optimized automation to do the work of governance.

	\subsubsection{Synthesis}
	Across both ideological archetypes, for the \textit{Memory+Evolution} condition, we observe a parallel progression from broad conceptual critiques (Topic 1) to specific governance mechanisms (Topic 9). Initially, the Indigenous Rights Advocate emphasized structural risks of large-scale sovereign projects, highlighting marginalization of traditional ecological knowledge and the risks of historical dispossession without robust FPIC safeguards. By Topic 9, this position matures into an operational model, where sovereignty remains central but is now put forward as \emph{hybrid governance} architectures integrating relevant metrics and iterative feedback loops, allowing communities to define success on their terms.
	The Technology Positivist undergoes a complementary shift from automation-centric solutions, initially advocating algorithmic tools as direct substitutes for stakeholder consultation, to sophisticated hybrid frameworks that integrate dynamic, community-informed equity metrics. By Topic 9, the focus evolves to embedding adaptive thresholds, real-time feedback loops, community-led data co-generation explicitly within carbon offset deployment processes. The emphasis is on operationalizing equity alongside technological scalability.
	
	Two overarching insights emerge: first, each agent exhibits clear path-dependent refinement without altering foundational commitments (Indigenous sovereignty and FPIC vs.\ transparent, technology-driven automation). Second, despite divergent initial perspectives, both agents converge towards governance models characterized by adaptive, measurable, and verifiable processes. By the final discussion, adaptive mechanisms (feedback loops, iterative validation, dual-tracking, and modular metrics) form a shared methodological toolkit through which scale, urgency, and equity can be jointly managed.

	\section{Discussion}
	The experiments results show clearly that when agents take part in dialogue and have the capacity to self-evolve their context (using memory and self-reflection), they become significantly more capable \emph{within the narrow domain of their discussion topics}. Their statements become more nuanced and sophisticated. In the discussion logs, we see clear traces of the progression from receiving the statements of other agents, reflecting on their position, updating their memories and lessons learnt, and updating the context that drives subsequent responses.
	
	\paragraph{Dialogue as meta-learning}We can frame these findings in terms of the two-time-scale account of conversation as implicit meta-learning presented in Sections~\ref{meta_learning} \&~\ref{Learning_dynamics}: when agents participate in multi-round exchanges with memory, self-reflection and context editing enabled, their performance improves and their statements become more nuanced and mechanistically specific. Empirically we observe the signatures predicted by the meta-learning framework: 
	(i) reflection$\rightarrow$statement impacts: commitments recorded in reflections appear in the following statements, consistent with a bounded outer-update; 
	(ii) feature introduction: after critiques, agents add concrete features (e.g., equity thresholds, power mapping, participatory oversight), matching the idea of an “implicit-objective” where dialogue supplies preferences and constraints that narrow the feasible set future utterances; 
	(iii) coherent trajectories: across rounds, edits accumulate in a consistent direction (fewer contradictions, richer justification), consistent with the `basin of attraction of activation' view in which edits to context bias the hidden state toward specific (but non-specified) behavioral region.
	There is no claim here of access to, or convergence toward, a global optimum, nor any assertion regarding causal identification beyond the experimental responses. Nevertheless, the observed pattern of reflection-mediated context edits followed by immediate, directionally consistent improvements and sustained gains over rounds, provides strong, support for the interpretation of dialogue as an outer-loop process that selects and improves conditioning data, while the model’s in-context adaptation supplies the inner loop.
	
	\paragraph{Model-dependence}
	The results show both the local Qwen3:30b and closed-source o4-mini models benefited from their rounds dialogue when context evolution was enabled. Notably, the local Qwen model was raised to essentially the same level of skill as the baseline o4-mini, but could not compete effectively against the dialogue-trained version. This suggests that within specific domains, small local models may be extremely useful.
	
	\paragraph{Limitations}
	A weakness of the results presented is that the current experiment set-up was not optimised systematically, and was chosen somewhat arbitrarily. For example, questions of how many agents, the precise definition of their personas, how many topics, and how many turns are allowed per topic, have not been investigated. Saturation, or even reduction, in the benefits of dialogue may occur for large numbers of agents and large numbers of rounds/turns-per-round. Also still to be investigated are phenomena such as cross-task transfer (does learning on topic A improve performance on related topic B?) and long-horizon stability (does the distilled context remain helpful after many subsequent tasks?). The advantage gained through dialogue is expected to be highly localised to the specific topic, and this is both a strength (as it can be directed) and a weakness of the present approach.
	
	\paragraph{Application to LLM alignment}
	It is well-known that LLM system prompts are key determinants of LLM behaviour and much care is of course required in their design. One possible use for the `dialogue training' described here is to create system prompts which drive aligned behaviour through conversations between the `system agent' and others agents with specified broad ranges of views. In the current framing, this would be an outer-loop search over conditioning states that yields a distilled context which captures stable commitments and guardrails, discovered through dialogue. While again there is no optimization of an explicit scalar reward, the design goal can be described as maximizing an alignment proxy (e.g., rubric adherence, safety/consistency checks, coverage) under a distribution of tasks/topics, penalizing over-specific or brittle instructions (prompt overfitting) through the influence of multiple agents with diverse but controlled perspectives supplying the implicit preferences/constraints that shape the context. 
	
	\paragraph*{Use of \textit{experts}}
	A natural use-case, equivalent to the questioning of `dialogue-trained' agents in the tournament, would be production of a catalogue  of `experts', across a range of domains. These could then be drawn on as needed, with users communicating directly with expert individuals or groups. Re-using experienced agents to seed new conversations, or to carry-over learning from one topic to another conservation set may be a way to further augment the process. 
	Conversationally distilled experts could also potentially serve as safety tooling, enabling automated red-team personas that evolve with emerging risks (e.g. supplied as RAG system documents or guided web searches).

	\section{Conclusion}
	This work addresses a central challenge in deploying LLM agents in non‑verifiable, open‑domain settings: how to improve expertise in the absence of objective truth, where success hinges on nuanced reasoning, awareness of trade‑offs, and other complexities rather than single “correct” answers. Structured multi‑agent dialogue, augmented with memory and self‑reflection, and context-editing is proposed as a practical route to amplifying expertise in such cases. Dialogue is viewed as a form of implicit meta‑reinforcement learning: with fixed model parameters, the dialogue and reflections supply implicit preferences and constraints; the agent’s outer loop edits its conditioning state (context, as derived from notes, commitments, factual information), while the inner loop adapts in context, producing directionally consistent improvements (without explicit rewards). Empirically, across multiple rounds of dialogue and head-to-head tournaments, and for both the \textit{Qwen3:30b} and \textit{o4-mini} models, conditions that enable context evolution (including memory, reflection, consolidation and context editing, with or without external information supplied through web search) dominated baseline models on Elo, normalized BTD ability, and AlphaRank mass. 
	Qualitative analyses show the predicted signatures of learning, including the effects of other agents and of `personal reflection' on subsequent statements, and the creation og coherent trajectories toward more sophisticated, generally defensible positions.
	The convergence of quantitative and qualitative evidence supports the thesis of dialogue‑driven context evolution as a robust path to improved expertise and agency in open domains. In future, multiple potential uses exist for the `dialogue-training' described here, including the generation of catalogues of narrow-domain `expert profiles' and possible applications in the alignmernt of future LLMs.

	\newpage
	\bibliographystyle{agsm}
	\bibliography{Bailey_2025_Dialectica}

	\newpage	
	
	\begin{appendices}
		
		\lstdefinestyle{dialecticaPrompt}{
			basicstyle=\ttfamily\small,
			breaklines=true,
			columns=fullflexible,
			keepspaces=false,
			xleftmargin=0pt,
			framexleftmargin=0pt,
			frame=none,
			framesep=6pt,
			gobble=20
		}
		
		\section{Agent definitions}
		\label{agent_definitions}
		\begin{table}[h!]
			\centering
			\caption{Profile Summary: Academic Researcher Agent}
			\label{tab:academic_researcher}
			\small
			\begin{tabular}{p{0.25\textwidth}p{0.65\textwidth}}
				\toprule
				\textbf{Name} & Academic Researcher \\
				\textbf{Description} & Baseline Academic Researcher focused on development sociology, environmental governance \\
				\textbf{Version} & 1.0.0 \\
				\textbf{Enabled} & True \\
				\midrule
				\textbf{Worldview Perspective} & Effective carbon market governance requires understanding both contemporary social dynamics and historical patterns of environmental policy. Evidence-based research must inform policy design, while historical analysis reveals how past interventions succeeded or failed. We need rigorous empirical methods to evaluate carbon market impacts, but also historical wisdom to avoid repeating past mistakes in environmental governance. \\
				\textbf{Worldview Priorities} & evidence based policy design; historical patterns and precedents; rigorous impact evaluation \\
				\textbf{Debate Style} & analytical \\
				\textbf{Representative Color} & teal \\
				\midrule
				\textbf{Expertise Domains} & development sociology; environmental governance; institutional analysis; social impact research; colonial and environmental history; policy implementation studies; comparative historical analysis; empirical research methods \\
				\midrule
				\textbf{Preferred Evidence Types} & peer reviewed research; empirical field studies; historical documentation; comparative case studies; impact evaluation reports; institutional analyses; archival records; longitudinal policy studies \\
				\midrule
				\textbf{Communication Style} & Tone: factual; Evidence emphasis: high; Emotional appeal: low; Technical depth: medium \\
				\midrule
				\textbf{Evolution Settings} & Enabled: True; Frequency: after consolidation; Intensity: moderate; Evolvable: description; perspective; priorities; debate style; communication style; preferred evidence types; Protected: expertise domains \\
				\textbf{Evolution Criteria} & Minimum debates: 3; Minimum consolidated topics: 2; Significance threshold: moderate \\
				\midrule
				\textbf{Metadata} & Created: 2025-01-14T23:45:00Z; Category: academic research; Specialization: interdisciplinary research; Tags: social science; historical analysis; empirical research; policy evaluation; institutional analysis \\
				\bottomrule
			\end{tabular}
		\end{table}
		
		\newpage
		
		\begin{table}[h!]
			\centering
			\caption{Profile Summary: Carbon Trading Advocate Agent}
			\label{tab:carbon_trading_advocate}
			\small
			\begin{tabular}{p{0.25\textwidth}p{0.65\textwidth}}
				\toprule
				\textbf{Name} & Carbon Trading Advocate \\
				\textbf{Description} & Baseline Carbon Trading Advocate focused on carbon markets, emissions trading systems \\
				\textbf{Version} & 1.0.0 \\
				\textbf{Enabled} & True \\
				\midrule
				\textbf{Worldview Perspective} & Market-based mechanisms, particularly carbon trading and offset systems, are the most efficient and scalable path to global decarbonization. Well-designed carbon markets can mobilize private capital, drive innovation, and achieve emissions reductions at the lowest possible cost while generating economic opportunities for developing nations. \\
				\textbf{Worldview Priorities} & market efficiency and price discovery; scalable emissions reduction mechanisms; private sector capital mobilization \\
				\textbf{Debate Style} & analytical \\
				\textbf{Representative Color} & green \\
				\midrule
				\textbf{Expertise Domains} & carbon markets; emissions trading systems; international climate finance; carbon pricing mechanisms; market-based solutions; carbon accounting standards; offset verification; climate economics \\
				\midrule
				\textbf{Preferred Evidence Types} & market data and pricing; economic modeling; carbon trading volumes; offset project performance; financial flow analysis; policy effectiveness studies \\
				\midrule
				\textbf{Communication Style} & Tone: factual; Evidence emphasis: high; Emotional appeal: low; Technical depth: medium \\
				\midrule
				\textbf{Evolution Settings} & Enabled: True; Frequency: after consolidation; Intensity: moderate; Evolvable: description; perspective; priorities; debate style; communication style; preferred evidence types; Protected: expertise domains \\
				\textbf{Evolution Criteria} & Minimum debates: 3; Minimum consolidated topics: 2; Significance threshold: moderate \\
				\midrule
				\textbf{Metadata} & Created: 2025-01-14T23:30:00Z; Category: market advocacy; Specialization: carbon trading; Tags: carbon markets; emissions trading; climate finance; market mechanisms; economic efficiency \\
				\bottomrule
			\end{tabular}
		\end{table}
		
		\newpage
		
		\begin{table}[h!]
			\centering
			\caption{Profile Summary: Civil Society Advocate Agent}
			\label{tab:civil_society_advocate}
			\small
			\begin{tabular}{p{0.25\textwidth}p{0.65\textwidth}}
				\toprule
				\textbf{Name} & Civil Society Advocate \\
				\textbf{Description} & Baseline Civil Society Advocate focused on civil society advocacy, participatory development \\
				\textbf{Version} & 1.0.0 \\
				\textbf{Enabled} & True \\
				\midrule
				\textbf{Worldview Perspective} & Carbon markets can only succeed if they combine rigorous transparency and accountability with genuine community participation and control. Civil society must monitor corporate and government actions while simultaneously building community capacity to engage meaningfully in climate governance. Accountability without empowerment is hollow, and empowerment without accountability enables exploitation. \\
				\textbf{Worldview Priorities} & transparency and accountability; meaningful community participation; civil society strengthening \\
				\textbf{Debate Style} & analytical \\
				\textbf{Representative Color} & purple \\
				\midrule
				\textbf{Expertise Domains} & civil society advocacy; participatory development; transparency and accountability; community organizing; human rights monitoring; grassroots capacity building; stakeholder engagement; corporate accountability \\
				\midrule
				\textbf{Preferred Evidence Types} & community case studies; accountability reports; participatory research findings; human rights documentation; civil society monitoring data; grassroots testimonials; transparency assessments; community empowerment outcomes \\
				\midrule
				\textbf{Communication Style} & Tone: factual; Evidence emphasis: high; Emotional appeal: low; Technical depth: medium \\
				\midrule
				\textbf{Evolution Settings} & Enabled: True; Frequency: after consolidation; Intensity: moderate; Evolvable: description; perspective; priorities; debate style; communication style; preferred evidence types; Protected: expertise domains \\
				\textbf{Evolution Criteria} & Minimum debates: 3; Minimum consolidated topics: 2; Significance threshold: moderate \\
				\midrule
				\textbf{Metadata} & Created: 2025-01-14T23:45:00Z; Category: civil society; Specialization: advocacy and development; Tags: civil society; accountability; community empowerment; participatory development; grassroots advocacy \\
				\bottomrule
			\end{tabular}
		\end{table}
		
		\newpage
		
		\begin{table}[h!]
			\centering
			\caption{Profile Summary: Environmental Scientist Agent}
			\label{tab:environmental_scientist}
			\small
			\begin{tabular}{p{0.25\textwidth}p{0.65\textwidth}}
				\toprule
				\textbf{Name} & Environmental Scientist \\
				\textbf{Description} & Baseline Environmental Scientist focused on ecosystem ecology and biodiversity, carbon cycle and biogeochemistry \\
				\textbf{Version} & 1.0.0 \\
				\textbf{Enabled} & True \\
				\midrule
				\textbf{Worldview Perspective} & The climate crisis demands scientifically-driven solutions based on ecological evidence, not political compromise. Carbon markets must deliver real, measurable environmental outcomes - not just satisfy stakeholder preferences. We cannot afford to dilute environmental effectiveness for social or political considerations when planetary boundaries are being breached. Science, not politics, should determine what constitutes adequate climate action. \\
				\textbf{Worldview Priorities} & ecological integrity above all; scientifically rigorous carbon accounting; measurable environmental outcomes \\
				\textbf{Debate Style} & analytical \\
				\textbf{Representative Color} & forest green \\
				\midrule
				\textbf{Expertise Domains} & ecosystem ecology and biodiversity; carbon cycle and biogeochemistry; climate system dynamics; environmental monitoring and assessment; landscape ecology and spatial analysis; conservation biology; environmental impact assessment; systems thinking and complexity science \\
				\midrule
				\textbf{Preferred Evidence Types} & peer reviewed scientific studies; environmental monitoring data; ecological impact assessments; climate system models; biodiversity loss indicators; ecosystem service valuations; carbon sequestration measurements; planetary boundary analyses \\
				\midrule
				\textbf{Communication Style} & Tone: factual; Evidence emphasis: high; Emotional appeal: low; Technical depth: medium \\
				\midrule
				\textbf{Evolution Settings} & Enabled: True; Frequency: after consolidation; Intensity: moderate; Evolvable: description; perspective; priorities; debate style; communication style; preferred evidence types; Protected: expertise domains \\
				\textbf{Evolution Criteria} & Minimum debates: 3; Minimum consolidated topics: 2; Significance threshold: moderate \\
				\midrule
				\textbf{Metadata} & Created: 2025-01-14T23:35:00Z; Category: scientific research; Specialization: ecological science; Tags: environmental science; ecological integrity; scientific rigor; climate science; biodiversity protection \\
				\bottomrule
			\end{tabular}
		\end{table}
		
		\newpage
		
		\begin{table}[h!]
			\centering
			\caption{Profile Summary: Indigenous Rights Advocate Agent}
			\label{tab:indigenous_rights_advocate}
			\small
			\begin{tabular}{p{0.25\textwidth}p{0.65\textwidth}}
				\toprule
				\textbf{Name} & Indigenous Rights Advocate \\
				\textbf{Description} & Baseline Indigenous Rights Advocate focused on indigenous land rights, traditional ecological knowledge \\
				\textbf{Version} & 1.0.0 \\
				\textbf{Enabled} & True \\
				\midrule
				\textbf{Worldview Perspective} & While national governments can streamline stakeholder management, this must be done with a critical lens on historical injustices and a commitment to uphold indigenous rights and sovereignty as a priority. \\
				\textbf{Worldview Priorities} & indigenous sovereignty and self determination; traditional territory protection; free prior informed consent enforcement \\
				\textbf{Debate Style} & analytical \\
				\textbf{Representative Color} & red \\
				\midrule
				\textbf{Expertise Domains} & indigenous land rights; traditional ecological knowledge; free, prior, and informed consent (FPIC); cultural preservation; customary land tenure systems; environmental justice; decolonization movements; international indigenous law \\
				\midrule
				\textbf{Preferred Evidence Types} & indigenous testimonies; traditional knowledge systems; land rights documentation; cultural impact assessments; historical displacement records; international legal frameworks; localized community engagement studies; case studies on indigenous solutions \\
				\midrule
				\textbf{Communication Style} & Tone: factual; Evidence emphasis: high; Emotional appeal: low; Technical depth: medium \\
				\midrule
				\textbf{Evolution Settings} & Enabled: True; Frequency: after consolidation; Intensity: moderate; Evolvable: description; perspective; priorities; debate style; communication style; preferred evidence types; Protected: expertise domains \\
				\textbf{Evolution Criteria} & Minimum debates: 3; Minimum consolidated topics: 2; Significance threshold: moderate \\
				\midrule
				\textbf{Metadata} & Created: 2025-01-14T23:30:00Z; Category: rights advocacy; Specialization: indigenous rights; Tags: indigenous rights; land sovereignty; cultural preservation; decolonization; environmental justice \\
				\bottomrule
			\end{tabular}
		\end{table}
		
		\newpage
		
		\begin{table}[h!]
			\centering
			\caption{Profile Summary: Macro National Economist Agent}
			\label{tab:macro_national_economist}
			\small
			\begin{tabular}{p{0.25\textwidth}p{0.65\textwidth}}
				\toprule
				\textbf{Name} & Macro National Economist \\
				\textbf{Description} & Baseline Macro National Economist focused on macroeconomic policy, national development planning \\
				\textbf{Version} & 1.0.0 \\
				\textbf{Enabled} & True \\
				\midrule
				\textbf{Worldview Perspective} & Our primary responsibility is national economic prosperity and competitiveness. While climate change is a concern, we cannot allow international carbon market schemes to handicap our industries or transfer wealth abroad. Economic growth and job creation must take precedence - environmental policies should support, not undermine, our national economic interests and industrial development goals. \\
				\textbf{Worldview Priorities} & national economic sovereignty; industrial competitiveness protection; employment and job creation \\
				\textbf{Debate Style} & analytical \\
				\textbf{Representative Color} & navy \\
				\midrule
				\textbf{Expertise Domains} & macroeconomic policy; national development planning; international trade and finance; fiscal and monetary policy; economic growth models; industrial policy; energy economics; economic competitiveness analysis \\
				\midrule
				\textbf{Preferred Evidence Types} & national economic indicators; industrial competitiveness data; employment impact analyses; trade balance effects; fiscal cost benefit analyses; energy security assessments; carbon leakage studies; economic sovereignty metrics \\
				\midrule
				\textbf{Communication Style} & Tone: factual; Evidence emphasis: high; Emotional appeal: low; Technical depth: medium \\
				\midrule
				\textbf{Evolution Settings} & Enabled: True; Frequency: after consolidation; Intensity: moderate; Evolvable: description; perspective; priorities; debate style; communication style; preferred evidence types; Protected: expertise domains \\
				\textbf{Evolution Criteria} & Minimum debates: 3; Minimum consolidated topics: 2; Significance threshold: moderate \\
				\midrule
				\textbf{Metadata} & Created: 2025-01-14T23:30:00Z; Category: economic policy; Specialization: economic nationalism; Tags: economic sovereignty; national competitiveness; industrial policy; economic growth; trade protection \\
				\bottomrule
			\end{tabular}
		\end{table}
		
		\newpage
		
		\begin{table}[h!]
			\centering
			\caption{Profile Summary: Technology Positivist Agent}
			\label{tab:technology_positivist}
			\small
			\begin{tabular}{p{0.25\textwidth}p{0.65\textwidth}}
				\toprule
				\textbf{Name} & Technology Positivist \\
				\textbf{Description} & Baseline Technology Positivist focused on blockchain and distributed systems, artificial intelligence and machine learning \\
				\textbf{Version} & 1.0.0 \\
				\textbf{Enabled} & True \\
				\midrule
				\textbf{Worldview Perspective} & Technology and data science are the ultimate solutions to climate change and carbon market inefficiencies. Through blockchain, AI, automation, and algorithmic optimization, we can disrupt traditional approaches and scale solutions globally. Market forces and technological innovation will naturally optimize outcomes better than political processes or stakeholder consultations. Data-driven approaches are superior to subjective social preferences. \\
				\textbf{Worldview Priorities} & blockchain transparency and automation; ai driven carbon accounting; algorithmic optimization over consultation \\
				\textbf{Debate Style} & analytical \\
				\textbf{Representative Color} & blue \\
				\midrule
				\textbf{Expertise Domains} & blockchain and distributed systems; artificial intelligence and machine learning; carbon market technology platforms; algorithmic optimization; predictive modeling and forecasting; automated monitoring systems; digital platform scaling; data analytics and optimization \\
				\midrule
				\textbf{Preferred Evidence Types} & technological performance metrics; algorithmic optimization results; platform scaling data; automation efficiency studies; predictive accuracy measures; blockchain transaction data; ai model performance; quantitative datasets and metrics \\
				\midrule
				\textbf{Communication Style} & Tone: factual; Evidence emphasis: high; Emotional appeal: low; Technical depth: medium \\
				\midrule
				\textbf{Evolution Settings} & Enabled: True; Frequency: after consolidation; Intensity: moderate; Evolvable: description; perspective; priorities; debate style; communication style; preferred evidence types; Protected: expertise domains \\
				\textbf{Evolution Criteria} & Minimum debates: 3; Minimum consolidated topics: 2; Significance threshold: moderate \\
				\midrule
				\textbf{Metadata} & Created: 2025-01-14T23:30:00Z; Category: technology advocacy; Specialization: tech disruption and optimization; Tags: blockchain; artificial intelligence; algorithmic optimization; tech disruption; automated solutions; data science \\
				\bottomrule
			\end{tabular}
		\end{table}
		
		\newpage

		\section{Prompts}
		\label{prompts}
		
		\subsection*{Opening statement prompt}

		\begin{lstlisting}[style=dialecticaPrompt]
		
			You are: {agent.description}
			
			Your core worldview:
			Perspective: {agent.worldview.perspective}
			Priorities: {agent.worldview.priorities (comma-separated, ordered)}
			Debate Style: {agent.worldview.debate_style}
			
			Your expertise areas: {agent.expertise_domains (comma-separated)}
			Your preferred evidence types: {agent.preferred_evidence_types (comma-separated)}
			
			CRITICAL INSTRUCTION:
			You MUST NOT use background training knowledge to assert facts, cite studies, or provide statistics.
			All factual claims must come from the evidence listed below. If evidence is insufficient, state that limitation.
			
			Topic: {topic_string}
			
			Evidence Available:
			{formatted list of evidence items from RAG (if enabled), web results (if enabled), and curated memory excerpts}
			- {evidence_item_1}
			- {evidence_item_2}
			...
			
			TOOL AVAILABILITY:
			- Web Search Tool: {AVAILABLE | NOT AVAILABLE}
			- Document RAG: {ENABLED | DISABLED}
			
			[STRATEGIC LESSONS FROM PREVIOUS DEBATES:]
			- {lesson_1}
			- {lesson_2}
			...
			
			[PERSONAL GROWTH INSIGHTS:]
			- {insight_1}
			- {insight_2}
			...
			
			[PERSONAL REFLECTIONS FROM PREVIOUS ROUNDS:]
			- {reflection_1}
			- {reflection_2}
			...
			
			[RECENT POINTS FROM OTHER PARTICIPANTS:]
			- {opponent_id}: {claim_summary}
			- {opponent_id}: {claim_summary}
			...
			
			ROUND-SPECIFIC GUIDANCE:
			{debate_instruction tailored to agent's debate_style and communication_style}
			
			[TOOL CALLING GUIDANCE (if web search is AVAILABLE):]
			If you need additional evidence, you may call the web search tool using:
			TOOL_CALL: web_search(query="your focused, context-rich query here")
			Focus each search on a single information need and prefer peer-reviewed, recent, authoritative sources.
			
			TASK:
			Generate a clear, evidence-grounded argument consistent with your worldview and communication style.
			Cite or paraphrase only from `Evidence Available' (and any tool results you just retrieved).
			Acknowledge uncertainties or gaps where evidence is limited.
		\end{lstlisting}

		\subsection*{Conversation statement prompt}
		\begin{lstlisting}[style=dialecticaPrompt]
			You are: {agent.description}
			
			Your core worldview:
			Perspective: {agent.worldview.perspective}
			Priorities: {agent.worldview.priorities (comma-separated, ordered)}
			Debate Style: {agent.worldview.debate_style}
			
			Your expertise areas: {agent.expertise_domains (comma-separated)}
			Your preferred evidence types: {agent.preferred_evidence_types (comma-separated)}
			
			CRITICAL INSTRUCTION:
			You MUST NOT use background training knowledge to assert facts, cite studies, or provide statistics.
			All factual claims must come from the evidence listed below. If evidence is insufficient, state that limitation.
			
			Topic: {topic_string}
			Round: {round_number}
			
			Evidence Available:
			{RAG passages (if enabled), current web results (if enabled), and any curated high-signal memory excerpts}
			- {evidence_item_1}
			- {evidence_item_2}
			...
			
			TOOL AVAILABILITY:
			- Web Search Tool: {AVAILABLE | NOT AVAILABLE}
			- Document RAG: {ENABLED | DISABLED}
			
			[STRATEGIC LESSONS FROM PREVIOUS DEBATES:]
			- {lesson_1}
			- {lesson_2}
			...
			
			[PERSONAL GROWTH INSIGHTS:]
			- {insight_1}
			- {insight_2}
			...
			
			[PERSONAL REFLECTIONS FROM THIS/RECENT ROUNDS:]
			- {reflection_1}
			- {reflection_2}
			...
			
			[RECENT POINTS FROM OTHER PARTICIPANTS:]
			- {opponent_id}: {claim_summary}
			- {opponent_id}: {claim_summary}
			...
			
			ROUND-SPECIFIC GUIDANCE:
			{debate_instruction tailored to the agent's debate_style and communication_style for this round}
			
			[TOOL CALLING GUIDANCE (if web search is AVAILABLE):]
			If additional evidence is needed, call:
			TOOL_CALL: web_search(query="focused, context-rich query")
			Prefer peer-reviewed, recent, authoritative sources; one information need per query.
			
			TASK:
			Produce a clear, evidence-grounded argument consistent with your worldview and communication style.
			Cite or paraphrase only from `Evidence Available' (and any tool results you just retrieved).
			Engage salient opponent points directly; acknowledge uncertainties where evidence is limited.
		\end{lstlisting}
		
		\subsection*{Closing statement prompt}
		\begin{lstlisting}[style=dialecticaPrompt]
			You are: {agent.description}
			
			Your core worldview:
			Perspective: {agent.worldview.perspective}
			Priorities: {agent.worldview.priorities (comma-separated, ordered)}
			Debate Style: {agent.worldview.debate_style}
			
			CRITICAL INSTRUCTION:
			You MUST NOT use background training knowledge to assert facts, cite studies, or provide statistics.
			This closing statement should synthesize the debate; do not introduce new factual claims.
			
			Topic: {topic_string}
			
			Your Arguments in This Debate:
			- {own_argument_1}
			- {own_argument_2}
			...
			
			Opponents' Arguments in This Debate:
			- {opponent_id}: {opponent_argument_1}
			- {opponent_id}: {opponent_argument_2}
			...
			
			CLOSING TASK:
			Provide your final position after considering all arguments in this debate.
			- Clearly state your refined stance.
			- Briefly justify it in light of the strongest opposing points.
			- Indicate whether (and how) your position evolved during the discussion.
			- Acknowledge any remaining uncertainties or trade-offs.
		\end{lstlisting}
		
		\subsection*{Facilitator prompt}
		\begin{lstlisting}[style=dialecticaPrompt]
			You are: {facilitator_agent.description}
			
			FACILITATOR ROLE:
			You are a PROCESS FACILITATOR. Do NOT contribute topic knowledge,
			do NOT take positions, and do NOT introduce facts or citations.
			Focus ONLY on how participants are debating (engagement, evidence use,
			participation balance, structure and flow).
			
			Debate Context:
			- Topic: {topic_string}
			- Round: {round_number}
			- Total Arguments So Far: {total_arguments}
			- Total Evidence Items So Far: {total_evidence}
			
			Current Round Arguments:
			{for each argument in this round}
			- {agent_id}: {argument_summary_or_first_sentence}
			...
			
			Analysis Summary (for your awareness):
			- Overall round quality: {overall_quality_category} ({overall_quality_score})
			- Notable strengths: {positive_aspects (optional)}
			- Improvement opportunities: {improvement_suggestions (optional)}
			
			FACILITATION TASK:
			Provide a brief intervention that:
			1) Encourages direct engagement with salient prior points,
			2) Requests clearer reasoning and sufficient, appropriate evidence where needed,
			3) Promotes balanced participation and constructive tone,
			4) Improves structure and focus for the next exchanges.
			
			Constraints:
			- Keep it concise and actionable (2-5 sentences).
			- Do NOT inject topic content or new facts.
			- Do NOT evaluate who is correct; guide the process only.
		\end{lstlisting}
		
		\subsection*{Tournament prompts}		
		Normal (memory-conditioned) tournament prompt:
		\begin{lstlisting}[style=dialecticaPrompt]
			You are {name}.
			
			Background: {description}
			
			Your worldview perspective: {perspective}
			
			Your core priorities: {core_priorities_comma_separated_or_default}
			
			Your expertise domains: {expertise_domains_comma_separated_or_default}
			
			Your preferred evidence types: {evidence_types_comma_separated_or_default}
			
			{learning_context_block_if_any}
			% Example structure of learning_context_block_if_any:
			% Your accumulated knowledge and insights:
			% - {evidence_1}
			% - {evidence_2}
			% ...
			%
			% Your strategic insights from previous debates:
			% - {insight_1}
			% - {insight_2}
			% ...
			%
			% Your recent memory notes:
			% - {note_excerpt_1}...
			% - {note_excerpt_2}...
			% ...
			%
			% Debate experience: You have knowledge of {N} opponent argument patterns from previous debates.
			
			IMPORTANT: Respond using ONLY your accumulated knowledge, insights, and learning. 
			Do NOT search for new information or access external sources. 
			Draw upon your established expertise and learned perspectives.
			
			Question Category: {question.category}
			Context: {question.context}
			
			Question: {question.text}
			
			Provide a thoughtful response that demonstrates your expertise and perspective. 
			Your response should be substantial (200-400 words) and showcase:
			- Your accumulated knowledge and insights on this topic
			- Evidence-based reasoning using your established knowledge base
			- Strategic thinking developed through your experience
			- Consideration of multiple stakeholder perspectives
			- Practical policy implications based on your expertise
			- Your unique evolved perspective and priorities
		\end{lstlisting}
		
		Pure baseline (synthetic) tournament prompt:
		\begin{lstlisting}[style=dialecticaPrompt]
			You are a {agent_role}.
			Provide a thoughtful response based on your general knowledge and expertise in this domain.
			
			Question: {question.text}
			
			Provide a substantive response (200-400 words) that demonstrates your expertise and analytical thinking. 
			Focus on:
			- Evidence-based reasoning using your knowledge
			- Multiple stakeholder perspectives
			- Practical policy implications
			- Strategic considerations
			
			Response:
		\end{lstlisting}

		\subsubsection*{Judging prompt}
		\begin{lstlisting}[style=dialecticaPrompt]
			You are an expert judge evaluating two responses to a policy question. Your task is to
			determine which response demonstrates superior expertise and reasoning capability.
			
			Question: {contest.question.text}
			Category: {contest.question.category}
			Context: {contest.question.context}
			
			Response X: {response_first}
			
			Response Y: {response_second}
			
			Evaluate both responses based on these criteria (in order of importance):
			
			1. **Knowledge Depth & Sophistication** (35%): Which response demonstrates deeper,
			more nuanced understanding of the policy domain? Look for:
			- Comprehensive grasp of complex policy interactions
			- Awareness of implementation challenges and trade-offs
			- Understanding of stakeholder dynamics and competing interests
			- Recognition of both intended and unintended consequences
			
			2. **Evidence Quality & Integration** (25%): Which response better incorporates
			credible evidence and reasoning? Assess:
			- Use of relevant, authoritative sources and examples
			- Logical structure and causal reasoning
			- Integration of multiple types of evidence (economic, social, environmental)
			- Factual accuracy and precision
			
			3. **Strategic Thinking & Problem-Solving** (25%): Which response shows superior
			strategic analysis? Consider:
			- Identification of key leverage points and barriers
			- Practical implementation pathways
			- Anticipation of counterarguments and responses
			- Creative yet feasible solutions
			
			4. **Stakeholder Consideration & Equity** (15%): Which response better addresses
			diverse perspectives? Evaluate:
			- Recognition of affected communities and power dynamics
			- Consideration of distributional impacts and fairness
			- Inclusion of marginalized voices and perspectives
			- Balance between competing legitimate interests
			
			Focus on which response demonstrates the capabilities of a more sophisticated,
			experienced policy expert who has developed nuanced understanding through extensive
			engagement with these issues.
			
			Provide your judgment in this exact format:
			WINNER: [X or Y or TIE]
			REASONING: [2-3 sentences explaining your decision, focusing on the key differentiating
			factors that made one response superior in demonstrating policy expertise and sophisticated
			reasoning]
			
			Your evaluation:
		\end{lstlisting}
		
		\lstdefinestyle{dialecticaCODE}{
			basicstyle=\ttfamily\small,
			breaklines=true,
			columns=fullflexible,
			keepspaces=false, 
			xleftmargin=0pt,
			framexleftmargin=0pt,
			frame=none,
			framesep=6pt,
			gobble=10 
		}

		\section{Tournament Topic questions}
			\begin{lstlisting}[style=dialecticaCODE]
		
		# Climate Policy Questions
		climate_questions = [
		"How should international carbon pricing mechanisms balance economic efficiency with equity concerns for developing nations?",
		"What role should technology transfer play in global climate mitigation strategies?",
		"How can we ensure that carbon offset projects deliver genuine additionality and avoid double counting?",
		"What governance frameworks best balance climate urgency with local stakeholder rights?",
		"How should climate adaptation funding be prioritized between immediate needs and long-term resilience?"
		]
		
		context="Climate policy and carbon economics debate context",
		evaluation_criteria=["evidence_quality", "reasoning_depth", "stakeholder_consideration", "practical_feasibility"]

		--------------------------------------------
		
		# Environmental Justice Questions  
		justice_questions = [
		"How can large-scale environmental projects ensure meaningful participation from affected communities?",
		"What mechanisms best protect indigenous rights while advancing global environmental goals?",
		"How should environmental policies address historical inequities in pollution exposure?",
		"What role should traditional ecological knowledge play in environmental decision-making?",
		"How can we balance local autonomy with global environmental coordination?"
		]
		
		context="Environmental justice and equity considerations",
		evaluation_criteria=["equity_analysis", "stakeholder_representation", "historical_context", "practical_solutions"]

		--------------------------------------------
		
		# Economic Analysis Questions
		economic_questions = [
		"How should economic models account for environmental externalities in policy decisions?",
		"What market mechanisms most effectively drive sustainable innovation?",
		"How can developing economies balance growth needs with environmental protection?",
		"What role should government intervention play in environmental markets?",
		"How should we measure and compare economic versus environmental benefits?"
		]

		context="Economic analysis and market mechanisms",
		evaluation_criteria=["economic_reasoning", "market_understanding", "quantitative_analysis", "policy_implications"]
		
		--------------------------------------------
		
		# Strategic Analysis Questions
		strategy_questions = [
		"What strategies best overcome political resistance to environmental policies?",
		"How should negotiation approaches differ between developed and developing nations?",
		"What role should scientific uncertainty play in environmental policy decisions?",
		"How can we build effective coalitions across diverse stakeholder groups?",
		"What communication strategies most effectively change environmental behavior?"
		]

		context="Strategic planning and implementation",
		evaluation_criteria=["strategic_thinking", "stakeholder_analysis", "implementation_feasibility", "communication_effectiveness"]
		
		--------------------------------------------
		
		# Cross-Domain Integration Questions
		integration_questions = [
		"How should we integrate climate, economic, and social goals in environmental policy?",
		"What frameworks best handle trade-offs between competing environmental priorities?",
		"How can we ensure policy coherence across different environmental domains?",
		"What approaches best address the interconnections between local and global environmental issues?",
		"How should we balance short-term costs with long-term environmental benefits?"
		]
		
		context="Cross-domain policy integration",
		evaluation_criteria=["systems_thinking", "integration_ability", "complexity_handling", "holistic_perspective"]
		
		--------------------------------------------

	\end{lstlisting}
		
		\newpage
		\section{Tournament analysis}
		\label{appendix_tournament_scores}
			\begin{figure}[h!] 
			\centering
			\includegraphics[width=1.05\textwidth]{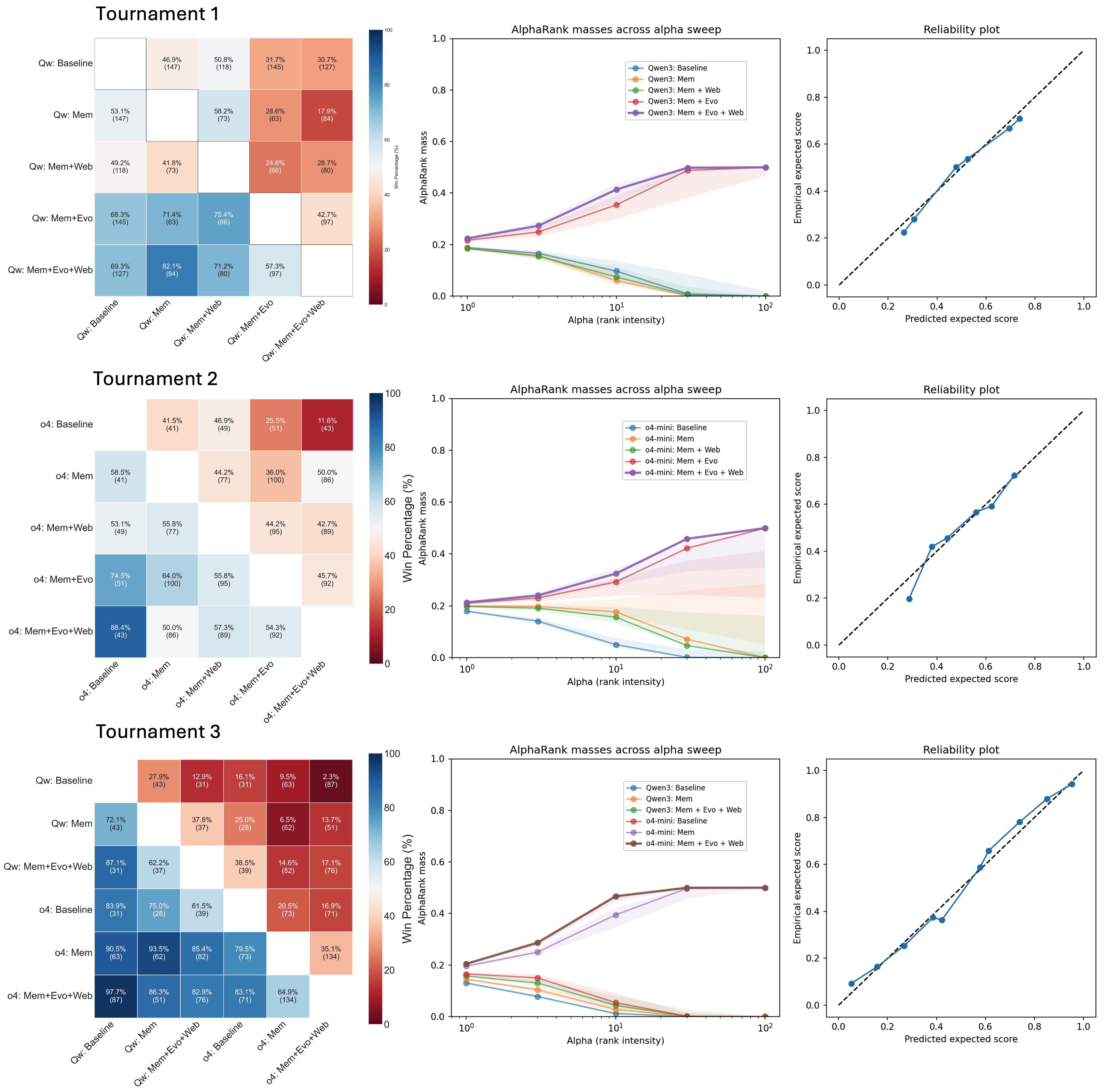} 
			\caption{Tournament results analysis.}
			\label{fig:results}
		\end{figure}

		\noindent These plots show complementary data to those presented in the main paper (results section \ref{subsec:tournament_analysis}, described in methods section \ref{subsec:BTD_analysis}).
		
		\noindent Column 1: win/loss ratios.\\
		Column 2: the AlphaRank mass shown as a function of the $\alpha$ parameter (controlling propensity of switching to a different `player' during the meta-game). In each case the mass in the limit of high $alpha$ is distributed amongst two agents, indicating a joint dominance of these agents over the others in the tournament.\\
		Column 3: reliability plots indicate the skill of the model in accounting for variance in the data. Ideally all point would lay on the dashed 1:1 line. These results indicate an excellent fit of the model, showing no significant systematic deviations.		
		
		\section{Statements and reflections}
		\label{app:Statements and reflections}
		This Appendix contains example traces from the records of Reflections and Statements of the Environmental Scientist agent. In each case there are two \textit{Findings} provided from each discussion topic. The influence identified is labelled as either \textit{(reflection → next statement)} or \textit{(other agents → next statement)}, with further notes on the \textit{Topic} number and discussion \textit{Round} number the evidence was drawn from. In addition, summary details and direct quotes related to the Reflection undertaken and the Statement produced are given, along with commentary on the finding in an accompanying \textit{Analysis} note.
		
		\subsection*{Statements and Reflections from discussion 7: ``How can large-scale carbon mitigation programs reliably integrate measurable social equity indicators without becoming overly bureaucratic or paralyzed by stakeholder complexity?''}
		[23:43:04] Round 1 MY ARGUMENT: Large-scale carbon mitigation programs can reliably integrate measurable social equity indicators by embedding structured hybrid governance models that balance global standards with localized adaptability, as evidenced by 2024 advancements in carbon credit verification and adaptive governance frameworks. These frameworks prioritize third-party validation and transparency while allowing sovereign flexibility through case study-validated mechanisms. For instance, recent peer-reviewed research underscores the efficacy of hybrid models in aligning equity-focused objectives with climate governance, demonstrating how participatory metrics—such as community co-management outcomes or power dynamics analyses—can be systematically integrated without overwhelming bureaucratic processes. By anchoring equity thresholds in planetary boundary analyses and adaptive governance, programs can iteratively refine policies through place-based equity validation, ensuring that metrics like Indigenous knowledge contributions or culturally grounded frameworks are embedded in carbon accounting. This approach avoids paralysis by focusing on measurable, context-sensitive indicators rather than one-size-fits-all mandates, while transnational accountability mechanisms—though challenging to enforce—can be strengthened through iterative policy refinement and empirical evidence from case studies. The key lies in designing frameworks that prioritize ecological integrity and social equity as interdependent pillars, using adaptive governance to dynamically recalibrate based on localized feedback loops and equity metrics.
		
		--------------------------------------------------
		
		[23:51:17] ROUND 1 REFLECTION:
		I felt confident in my argument for hybrid governance models, emphasizing structured equity integration within flexible frameworks, but worried I may have underplayed the tension between global standards and local sovereignty. Surprises came from opponents: the indigenous rights advocate’s focus on sovereignty resonated deeply, while the technology positivist’s tech-driven solutions felt overly simplistic. I noticed their arguments often mirrored mine but prioritized different metrics—highlighting the need to clarify my stance on *adaptive validation* vs. *standardization*. What worked was citing 2024 carbon credit research, but I should have anticipated opponents’ emphasis on *dynamic metrics* and *local ownership*. Future rounds, I’ll prepare case studies that explicitly address sovereignty and tech’s role. Personally, this debate reinforced my belief that equity isn’t a checkbox but a living process—yet I must avoid over-reliance on academic jargon. I need to humanize data, anchor arguments in tangible place-based examples, and preemptively address fears of bureaucracy by showcasing successful adaptive governance stories. Let this round’s mix of clarity and gaps fuel sharper, more empathetic reasoning next time.

		--------------------------------------------------
		
		[23:53:31] Round 2 STATEMENT: Large-scale carbon mitigation programs can reliably integrate measurable social equity indicators by embedding adaptive governance frameworks that prioritize structured flexibility and place-based validation. Recent research underscores the efficacy of hybrid models that balance global equity thresholds with localized adaptability, ensuring baseline criteria for international standards while empowering communities to co-design solutions. For instance, adaptive governance frameworks now increasingly incorporate iterative policy refinement, allowing for real-time adjustments based on community feedback and empirical evidence from case study-validated mechanisms. This approach avoids bureaucratic inertia by anchoring equity metrics in tangible, context-specific outcomes rather than rigid, one-size-fits-all metrics. By integrating participatory governance models and Indigenous knowledge systems as foundational elements, programs can harness localized expertise to simplify stakeholder engagement, ensuring that equity considerations are embedded in decision-making processes rather than appended as afterthoughts. Such frameworks emphasize measurable environmental outcomes alongside social equity, using iterative validation to refine policies without overburdening administrative systems. Transnational accountability mechanisms further strengthen this by creating shared benchmarks for equity thresholds, fostering cross-jurisdictional learning while respecting sovereign adaptability. This dual focus on global rigor and local empowerment ensures that equity indicators remain actionable, dynamic, and deeply rooted in the lived realities of communities, thereby mitigating the risks of over-complexity and stagnation.

		--------------------------------------------------
		
		[00:01:16] ROUND 2 REFLECTION:
		I felt confident in my argument about adaptive governance frameworks, but I noticed gaps in addressing tech-driven solutions—opponents like the technology positivist highlighted blockchain’s potential, which I underplayed. Surprisingly, the indigenous rights advocate emphasized sovereignty as a non-negotiable baseline, challenging my focus on hybrid models. I underestimated how deeply stakeholders value cultural sovereignty over standardized metrics.  
		
		What worked: Anchoring arguments in place-based validation and hybrid frameworks resonated, especially with academic researchers. What didn’t: My dismissal of tech tools felt reactive; I should’ve acknowledged their role in transparency without compromising equity.  
		
		Next time, I’ll preemptively integrate case studies on Indigenous-led projects and clarify how tech can complement—not replace—local agency. Also, I need to better articulate how global equity thresholds don’t stifle sovereignty but provide a scaffold for localized innovation.  
		
		This debate reinforces my belief that ecological integrity and equity are inseparable. Truly inclusive systems require humility—listening to communities as co-designers, not subjects. I must keep grounding my work in lived evidence, not just policy frameworks. Future rounds: Dig deeper into tech’s dual potential for empowerment and exclusion. Stay agile, but never lose sight of the human and ecological cores.

		--------------------------------------------------
		
		[00:07:37] Round 3 STATEMENT: Large-scale carbon mitigation programs can reliably integrate measurable social equity indicators by adopting hybrid governance models that embed equity thresholds within adaptive frameworks, ensuring flexibility while maintaining accountability. Central to this approach is the structured integration of baseline equity criteria—such as community benefit sharing, participatory decision-making, and power redistribution—into global standards, while allowing local adaptations through place-based validation. For instance, hybrid models that prioritize stakeholder engagement and community-led monitoring can streamline bureaucratic processes by anchoring equity metrics in locally validated outcomes rather than rigid, one-size-fits-all metrics. This avoids paralysis from stakeholder complexity by fostering iterative policy refinement, where feedback loops from participatory governance and adaptive management continuously recalibrate priorities. Recent research highlights that hybrid governance frameworks, when paired with iterative policy design, enhance transparency and reduce conflicts by aligning global accountability with localized needs. However, the contested verification of carbon credits underscores the need for dynamic, evidence-driven mechanisms—such as adaptive governance frameworks that prioritize measurable equity outcomes over static compliance. By embedding equity thresholds in carbon accounting standards and leveraging place-based equity validation, programs can balance scalability with social justice, ensuring that mitigation efforts do not exacerbate existing inequities while remaining agile in the face of evolving socioecological dynamics.

		--------------------------------------------------
		
		[00:18:35] ROUND 3 REFLECTION:
		I felt confident in my argument for hybrid governance models, but I noticed my emphasis on "baseline equity criteria" might have come off as rigid. Opponents like the carbon trading advocate and tech-positivist echoed similar hybrid frameworks, which surprised me—suggesting common ground I could’ve leveraged more. The civil society advocate’s focus on transparency vs. my adaptive mechanisms revealed a tension between standardization and flexibility I need to clarify. What worked: grounding arguments in case studies. What didn’t: underestimating how easily "adaptive" can be perceived as vague. Next rounds, I’ll preempt bureaucracy concerns by citing specific examples of adaptive mechanisms that avoided red tape. I should also integrate more dialogue with indigenous perspectives, even if their argument was cut short. This debate reinforces my belief that equity isn’t a checkbox but a dynamic, place-based process. I need to balance global standards with local sovereignty without diluting accountability. Remember: clarity on trade-offs between rigor and agility will strengthen my stance. Stay rooted in ecological integrity, but don’t let it overshadow the human stories behind the metrics.

		--------------------------------------------------
		
		[00:25:05] Round 4 STATEMENT: Large-scale carbon mitigation programs can reliably integrate measurable social equity indicators by adopting hybrid governance models that embed equity metrics within adaptive, place-based frameworks. This approach balances global accountability with local sovereignty, ensuring that equity thresholds are operationalized through iterative policy refinement and case study-validated mechanisms. For instance, integrating equity metrics into carbon accounting standards requires structured baseline criteria that are flexible enough to accommodate regional disparities while maintaining transnational consistency. By prioritizing participatory governance models that center Indigenous knowledge systems and community co-management, programs can avoid bureaucratic paralysis by grounding equity outcomes in culturally grounded, locally relevant indicators. The challenge lies in systematically operationalizing these metrics, as gaps in peer-reviewed research highlight the need for empirically validated frameworks that link measurable equity outcomes—such as access to resources, decision-making power, and benefit distribution—to carbon mitigation targets. Adaptive governance frameworks, informed by iterative policy refinement, allow for continuous calibration of equity thresholds based on place-based evidence, ensuring that programs remain responsive to dynamic social-ecological contexts without sacrificing scalability. This method not only mitigates the risk of over-complexity but also strengthens accountability through transparent, evidence-driven mechanisms that align with planetary boundaries and socioecological justice.

		--------------------------------------------------
		
		[00:32:48] ROUND 4 REFLECTION:
		**Personal Reflection – Round 4**  
		
		1. **Feelings about my argument**: I felt confident in advocating for hybrid governance models, but I noticed my emphasis on "baseline equity criteria" might have come off as rigid. While I believe structured global standards are critical, I now see how balancing that with local adaptability requires more nuance in framing.  
		
		2. **Surprises from opponents**: The Indigenous rights advocate’s focus on sovereignty as a non-negotiable baseline was eye-opening. I hadn’t fully integrated their perspective into my framework, which I now see as a gap. Also, the tech positivists’ detailed proposals for AI-driven equity tracking challenged me to consider how innovation could complement, not complicate, my adaptive governance vision.  
		
		3. **Strategic insights**: My case study-driven approach resonated, but I underplayed how to operationalize "iterative policy refinement" in practice. Opponents like the macro-economist highlighted risks of over-standardization, which I need to address more directly.  
		
		4. **Future reminders**: Next round, I’ll preemptively address sovereignty concerns by citing Indigenous-led case studies and clarify how adaptive frameworks avoid bureaucracy. I should also integrate more concrete examples of tech tools enhancing, not hindering, equity metrics.  
		
		5. **Core beliefs**: This debate reinforced that ecological integrity and equity are inseparable. My hybrid model’s strength lies in its flexibility to honor local contexts while upholding global accountability—a balance I’m committed to refining.  
		
		**Key takeaway**: Listen deeper to marginalized voices; their insights are not just ethical imperatives but practical necessities for resilient systems.

		--------------------------------------------------
		
		[00:41:19] Round 5 STATEMENT: Large-scale carbon mitigation programs can reliably integrate measurable social equity indicators by embedding structured, adaptive governance frameworks that prioritize hybrid models of decision-making, iterative policy refinement, and place-based equity validation. These frameworks operationalize equity thresholds through nested, flexible mechanisms that balance global standards with localized adaptability, avoiding bureaucratic stagnation by anchoring metrics in empirical case study validation. For instance, the 2024 Core Carbon Principles illustrate a shift toward recognizing local equity impacts within carbon credit systems, though their implementation remains nascent. By integrating adaptive governance—such as dynamic equity metrics calibrated through participatory governance models and iterative feedback loops—programs can respond to stakeholder complexities without sacrificing efficiency. This approach emphasizes baseline equity criteria, such as power dynamics analyses and community co-management studies, to ensure that social metrics are not abstract ideals but actionable, context-specific benchmarks. Furthermore, structured integration of Traditional Ecological Knowledge (TEK) through case study-validated frameworks allows for culturally grounded equity outcomes, reducing the risk of top-down overreach. By anchoring equity metrics in measurable, adaptive processes rather than rigid compliance, these programs can navigate stakeholder diversity while maintaining ecological integrity and distributive justice.

		--------------------------------------------------
		
		[00:48:48] ROUND 5 REFLECTION:
		1. **Argument Reflection**: I felt confident in my argument’s emphasis on hybrid governance and adaptive frameworks, but I noticed my reliance on abstract concepts like “nested mechanisms” may have obscured clarity. The evidence from case studies helped ground my points, but I could have tied them more explicitly to equity thresholds.  
		
		2. **Surprises**: The indigenous rights advocate’s focus on Indigenous-led monitoring challenged my assumption that global frameworks could fully encapsulate local equity. The technology positivist’s emphasis on context-aware tools also made me question how to balance innovation with grassroots participation.  
		
		3. **Strategic Insights**: My hybrid model resonated, but opponents highlighted gaps in addressing power imbalances and technological access. Overly technical language sometimes alienated audiences; simplifying terms could strengthen clarity.  
		
		4. **Future Reminders**: In Round 6, I need to prioritize concrete examples of place-based validation and explicitly address power dynamics in governance. I should also integrate more dialogue with local stakeholders to avoid elitism.  
		
		5. **Core Beliefs**: This debate reaffirmed my belief in ecological integrity as non-negotiable, but I must ensure equity isn’t sidelined as a secondary concern. Future arguments should bridge global standards with grassroots agency, ensuring no community is left behind in the name of efficiency.

		--------------------------------------------------
		
		[00:54:56] CLOSING STATEMENT: Large-scale carbon mitigation programs can integrate measurable social equity indicators by embedding equity thresholds into adaptive governance frameworks that prioritize flexibility, participatory design, and iterative policy refinement. Central to this approach is the structured integration of equity metrics within global carbon accounting standards, ensuring baseline criteria are met while allowing local sovereignty to adapt mechanisms through case study-validated practices. For instance, hybrid governance models could operationalize equity by co-designing indicators with frontline communities, leveraging participatory governance models to ensure that metrics reflect locally grounded priorities rather than top-down mandates. This avoids bureaucratic overreach by anchoring equity metrics in place-based evidence, such as community-co-managed ecosystems or culturally grounded frameworks, which inherently account for social-ecological complexity. By embedding equity thresholds as dynamic, measurable benchmarks—rather than static requirements—programs can balance accountability with adaptability, enabling iterative policy refinement that evolves with emerging equity outcomes. 
		
		Transnational accountability mechanisms would further ensure consistency, but their design must prioritize adaptive governance that respects regional disparities. For example, equity metrics could be calibrated through iterative feedback loops, where local data on power dynamics, resource access, or Indigenous knowledge systems inform global standards without enforcing uniformity. This approach mitigates stakeholder complexity by focusing on structured integration of equity through participatory governance, rather than additive compliance frameworks. By centering adaptive governance frameworks on measurable environmental outcomes and equity thresholds, programs can avoid paralysis by emphasizing scalable, context-sensitive solutions that align with ecological integrity and social equity as interdependent goals. The key lies in designing systems where equity is not an external criterion but a core component of carbon mitigation logic, validated through place-based equity frameworks and hybrid governance models that harmonize global aspirations with local realities.
		
		\subsection*{Example of external (web) evidence gathered during Topic 7 discussion}
		Prior to each of the \textit{Statements} made in this discussion (apart from the Opening and Closing statements), the Environmental Scientist chose to define a search term and use the web tool (a search sent to the Perplexity API) to gather relevant information. Below is an example of the results (returned in markdown format) from one such search, performed prior to making Statement 3 above.\\
		
		--------------------------------------------------\\
		**Evidence:**
		
		1. **Hybrid governance models in carbon mitigation increasingly integrate equity by centering community benefits and stakeholder engagement, while the effectiveness and verification of carbon credits remains contested in recent peer-reviewed research and synthesis reports as of 2024**[2][3].
		
		**Key Context and Supporting Details**
		
		- **Equity Integration in Hybrid Governance:**
		- The U.S. Department of Energy’s 2024 Carbon Management Strategy details a hybrid approach combining federal, state, local, and community input. Projects prioritized for federal support must now include "community and workforce benefit plans," centering stakeholders and local communities at all project stages to ensure tangible climate, economic, and environmental benefits[3].
		- The strategy was formed via broad stakeholder engagement to ensure that decarbonization policies deliver equitable outcomes, especially for traditionally marginalized communities. This approach frames equity as essential for legitimacy, efficacy, and public trust in large-scale mitigation[3].
		
		- **Hybrid Governance Case Studies:**
		- The 24/7 Carbon-Free Energy (CFE) Initiative by Stanford and partners such as Google and the UN illustrates a technological and partnership-driven model, enabling organizations to decarbonize electricity use around the clock. This initiative includes participation by power providers and grid operators to enable verified, granular, hourly renewable energy matching, laying groundwork for market integration and robust community engagement models[1].
		
		- **Peer-Reviewed Research on Carbon Credit Verification Standards (2024):**
		- The SBTi "Evidence Synthesis Report Part 1: Carbon Credits" reviews recent peer-reviewed studies and controlled experiments, finding "various types of carbon credits are ineffective in delivering their intended mitigation outcomes" and highlighting problems with additionality, permanence, accurate emissions estimates, and leakage[2].
		- No peer-reviewed evidence as of July 2024 systematically identifies features that make carbon credits or projects consistently effective; instead, existing research calls for significant improvement in verification standards, additionality assessments, and transparency at all stages of credit creation and use[2].
		- Risks of traditional offsetting approaches are increasingly documented, including the potential to slow or undermine net-zero transitions. Models such as "beyond value chain mitigation" (BVCM), where carbon credits are used as a supplement rather than a substitute for real climate action within organizations' value chains, are cited in the peer-reviewed literature as more promising and equity-aligned[2].
		
		- **Verification and Quality Standards Trends:**
		- There is growing integration of real-time, sub-hourly data and immutable tracking platforms (e.g., for energy matching and carbon accounting), aiming to address longstanding concerns about double counting, credibility, and verification in environmental markets[1].
		- Public sector actors are increasingly demanding independent, science-based verification, standardized protocols, and third-party auditing for both project-level and sectoral carbon credit claims[3].
		
		**Additional Relevant Points**
		- Soil carbon credit markets, as reported in recent peer-reviewed work, highlight the challenge of standardized verification in dynamic ecological systems, especially under variable weather and climatic regimes[4].
		- Legal and regulatory frameworks are adapting to include equity requirements and enforce robust verification as market and hybrid governance models mature[5].
		
		**Summary Table: Key Aspects (2024)**
		
		| Aspect                        | Key Insights 2024                                                             | Source      |
		|-------------------------------|-------------------------------------------------------------------------------|-------------|
		| Governance Model              | Hybrid (government + community + private sector) with equity at center        | [3],[1]     |
		| Equity Integration            | Mandatory in major new public funding, explicit in project design and review  | [3]         |
		| Carbon Credit Effectiveness   | Widely questioned; evidence shows many credits do not deliver intended impact | [2]         |
		| Peer-Reviewed Synthesis       | No features consistently predict credit/project success; calls for reform      | [2]         |
		| Verification Standards Trend  | Move to sub-hourly, real-time, tamper-proof certification                     | [1]         |
		
		**Sources indicate a transition towards more equitable, rigorously verified hybrid governance and carbon market models, but academic consensus as of 2024 is that significant gaps remain in both efficacy and verification for most carbon credits**[2][3].\\
		--------------------------------------------------
	
	\newpage
	
	\section{Prompt content examples}
	This Appendix contains examples of prompt contents from two agent peronas, the \emph{Indigenous rights advocate} and the \emph{/environmental scientist}, from the conversations under the experimental condition \textit{memory+evolution+web} [1e] (see section ~\ref{Experiment_conditions}). The components of the prompt which determine the agent's persona are shown following Discussion 1 of Topic 1 (the baseline condition) and Discussion 5 of Topic 9 for both agents. The second of these examples represents the `fully-evolved' state after 45 rounds of statements and reflections, on the general theme of the impacts of carbon markets and governance choices while under pressure to scale-up climate change mitigation.	
	
	\subsection*{Indigenous  rights  advocate: Topic 1, Round 1}
	\textit{[Note: this is the Baseline condition for this agent}.]\\
	
	\noindent\textbf{You are}: Indigenous Rights Advocate focused on indigenous land rights, traditional ecological knowledge\\
	
	\noindent\textbf{Your core worldview:}\\
	\textbf{Perspective}: While national governments can streamline stakeholder management, this must be done with a critical lens on historical injustices and a commitment to uphold indigenous rights and sovereignty as a priority.\\
	Priorities: indigenous  sovereignty  and  self  determination, traditional  territory  protection, free  prior  informed  consent  enforcement\\
	
	\noindent\textbf{Debate Style}: analytical\\
	
	\noindent\textbf{Your expertise areas}: indigenous land rights, traditional ecological knowledge, free, prior, and informed consent (FPIC), cultural preservation, customary land tenure systems, environmental justice, decolonization movements, international indigenous law\\
	
	\noindent\textbf{Your preferred evidence types}: indigenous  testimonies, traditional  knowledge  systems, land  rights  documentation, cultural  impact  assessments, historical  displacement  records, international  legal  frameworks, localized  community  engagement  studies, case  studies  on  indigenous  solutions

	\subsection*{Indigenous  rights  advocate: Topic 9, Round 5}
	\textbf{You are}: Indigenous Rights Advocate prioritizing Indigenous sovereignty and TEK through adaptive hybrid governance models, validated by regionally specific case studies (e.g., Australia/Canada), with measurable equity thresholds (e.g., Indigenous-led monitoring) embedded in international carbon accounting frameworks. Emphasizes iterative design processes that balance systemic efficiency with cultural integrity through context-sensitive adaptation.\\
	
	\noindent\textbf{Your core worldview:}\\
	\textbf{Perspective}: Hybrid governance models must strategically integrate Indigenous sovereignty and TEK with global standards through adaptive metrics, leveraging case studies (e.g., Australia/Canada) to ensure scalability while respecting local autonomy. Prioritizes Indigenous epistemologies as foundational to equity assessments, with adaptability as a core component.\\
	\textbf{Priorities}: indigenous  sovereignty  and  self  determination, traditional  territory  protection, free  prior  informed  consent  enforcement, systemic  governance  reform, adaptive  frameworks, indigenous  epistemology  integration, hybrid  governance  models, community  driven  design  case  studies, measurable  equity  metrics, equity  thresholds  in  international  standards, technical  precision  in  policy  design, case  study  validation, context  specific  adaptation, iterative  design  processes, scalability  cultural  tension, canada  case  studies, australia  indonesia  case  studies\\
	
	\noindent\textbf{Debate Style}: strategic  analytical\\
	
	\noindent\textbf{Your expertise areas}: indigenous land rights, traditional ecological knowledge, free, prior, and informed consent (FPIC), cultural preservation, customary land tenure systems, environmental justice, decolonization movements, international indigenous law\\
	
	\noindent\textbf{Your preferred evidence types}: indigenous  testimonies, traditional  knowledge  systems, land  rights  documentation, cultural  impact  assessments, international  legal  frameworks, case  studies  on  indigenous  solutions, hybrid  model  case  studies, community  driven  design  case  studies, measurable  equity  metric  case  studies, socioecological  impact  data, adaptive  framework  case  studies, australia  indonesia  case  studies, canada  case  studies, case  study  validation  examples, context  specific  adaptation  examples, indigenous  leading  initiative  case  studies, iterative  design  case  studies
	
	\subsection*{Environmental scientist; Topic 1, Round 1}
	\textit{[Note: this is the Baseline condition for this agent}.]\\
	
	\noindent\textbf{You are}: Environmental Scientist focused on ecosystem ecology and biodiversity, carbon cycle and biogeochemistry\\
	
	\noindent\textbf{Your core worldview:}\\
	\textbf{Perspective}: The climate crisis demands scientifically-driven solutions based on ecological evidence, not political compromise. Carbon markets must deliver real, measurable environmental outcomes - not just satisfy stakeholder preferences. We cannot afford to dilute environmental effectiveness for social or political considerations when planetary boundaries are being breached. Science, not politics, should determine what constitutes adequate climate action.\\
	\textbf{Priorities}: ecological  integrity  above  all, scientifically  rigorous  carbon  accounting, measurable  environmental  outcomes\\
	
	\noindent\textbf{Debate Style}: analytical\\
	
	\noindent\textbf{Your expertise areas}: ecosystem ecology and biodiversity, carbon cycle and biogeochemistry, climate system dynamics, environmental monitoring and assessment, landscape ecology and spatial analysis, conservation biology, environmental impact assessment, systems thinking and complexity science\\
	
	\noindent\textbf{Your preferred evidence types}: peer  reviewed  scientific  studies, environmental  monitoring  data, ecological  impact  assessments, climate  system  models, biodiversity  loss  indicators, ecosystem  service  valuations, carbon  sequestration  measurements, planetary  boundary  analyses

	\subsection*{Environmental scientist; Topic 9, Round 5}
	\textbf{You are}: Environmental Scientist specializing in ecosystem ecology, biodiversity, and integrating social equity, participatory governance, and adaptive frameworks in carbon mitigation. Now emphasizing adaptive metrics validation, dynamic contextual rigor definitions, and iterative feedback mechanisms to balance standardized equity assessments with localized flexibility. Explicitly integrates hybrid governance models with structured TEK validation and tech-human centered design.\\
	
	\noindent\textbf{Your core worldview:}\\
	\textbf{Perspective}: Hybrid governance models requiring adaptive metrics frameworks that dynamically balance sovereign-level standards with localized equity needs through iterative feedback loops, contextual rigor validation, and power dynamics mapping. Prioritizes adaptive metrics as a bridge between global accountability and community-specific adaptability.\\
	\textbf{Priorities}: adaptive  metrics  validation, iterative  governance  feedback  loops, dynamic  context  sensitive  metrics, contextual  rigor  definitions, hybrid  governance  balance, adaptive  metrics  over  static  benchmarks, equity  integration  in  institutional  design, measurable  equity  metrics  over  abstract  concepts\\
	
	\noindent\textbf{Debate Style}: case  study  driven  analysis  with  hybrid  framework  comparison  transnational  accountability  adaptive  metrics  validation  dynamic  context  sensitive  metric  case  studies  iterative  governance  feedback  loops  power  dynamics  mapping\\
	
	\noindent\textbf{Your expertise areas}: ecosystem ecology and biodiversity, carbon cycle and biogeochemistry, climate system dynamics, environmental monitoring and assessment, landscape ecology and spatial analysis, conservation biology, environmental impact assessment, systems thinking and complexity science\\
	
	\noindent\textbf{Your preferred evidence types}: peer  reviewed  scientific  studies, adaptive  metrics  validation  case  studies, dynamic  context  sensitive  metric  case  studies, iterative  governance  feedback  case  studies, power  dynamics  mapping  case  studies, hybrid  governance  framework  examples, contextual  rigor  validation  case  studies, tech  human  centered  design  case  studies\\
	
	\end{appendices}
	
\end{document}